\pdfoutput=1

\documentclass[11pt]{article}

\usepackage[final]{acl}

\usepackage{times}
\usepackage{latexsym}

\usepackage[T1]{fontenc}

\usepackage[utf8]{inputenc}

\usepackage{microtype}

\usepackage{inconsolata}

\usepackage{graphicx}
\usepackage{amsmath}
\usepackage{latexsym}
\usepackage{amssymb}
\usepackage{adjustbox}
\usepackage{kotex}
\usepackage{booktabs}
\usepackage{multirow}
\usepackage{subcaption}
\usepackage{array}
\newcolumntype{M}[1]{>{\raggedright\arraybackslash}m{#1}} 

\usepackage[most]{tcolorbox}
\usepackage{enumitem}
\usepackage{courier} 
\usepackage[utf8]{inputenc}
\usepackage{lipsum}

\usepackage{colortbl}
\usepackage{hyperref}

\newtcolorbox{instructionbox}[1]{
  colback=gray!10,
  colframe=gray!60!black!40,
  title=\textbf{#1},
  fonttitle=\bfseries\large,
  coltitle=black,
  boxrule=1pt,
  arc=5pt,
  top=2mm,
  bottom=2mm,
  left=2mm,
  right=2mm,
  width=0.49\textwidth,
  sharp corners=south,
  before upper={\ttfamily},
  breakable
}

\title{ScholarBench: A Bilingual Benchmark for Abstraction, \\ Comprehension, and Reasoning Evaluation in Academic Contexts}

\author{
    Dongwon Noh\textsuperscript{1}\thanks{Equal contribution}, Donghyeok Koh\textsuperscript{1}\footnotemark[1], Junghun Yuk\textsuperscript{2}\footnotemark[1], 
    Gyuwan Kim\textsuperscript{3}\footnotemark[1] \\
    \textbf{
        Jaeyong Lee\textsuperscript{4},
        KyungTae Lim\textsuperscript{2}\footnotemark[2],
        Cheoneum Park\textsuperscript{1}\thanks{Corresponding authors.}
        } \\ \\
    \textsuperscript{1}HBNU,
    \textsuperscript{2}KAIST,
    \textsuperscript{3}UCSB,
    \textsuperscript{4}KISTI
}

\begin{document}
\maketitle
\begin{abstract}
Prior benchmarks for evaluating the domain-specific knowledge of large language models (LLMs) lack the scalability to handle complex academic tasks.
To address this, we introduce \texttt{ScholarBench}, a benchmark centered on deep expert knowledge and complex academic problem-solving, which evaluates the academic reasoning ability of LLMs and is constructed through a three-step process.
\texttt{ScholarBench} targets more specialized and logically complex contexts derived from academic literature, encompassing five distinct problem types. Unlike prior benchmarks, \texttt{ScholarBench} evaluates the abstraction, comprehension, and reasoning capabilities of LLMs across eight distinct research domains. To ensure high-quality evaluation data, we define category-specific example attributes and design questions that are aligned with the characteristic research methodologies and discourse structures of each domain. Additionally, this benchmark operates as an English-Korean bilingual dataset, facilitating simultaneous evaluation for linguistic capabilities of LLMs in both languages. The benchmark comprises 5,031 examples in Korean and 5,309 in English, with even state-of-the-art models like o3-mini achieving an average evaluation score of only 0.543, demonstrating the challenging nature of this benchmark. Evaluation code and dataset are available at \url{https://github.com/hbnu-kilab/ScholarBench}
\end{abstract}

\section{Introduction}
\noindent
The emergence and application of large language models (LLMs)~\cite{openai2024gpt4technicalreport, touvron2023llamaopenefficientfoundation, gemmateam2024gemma2improvingopen} has significantly advanced performance across diverse natural language processing tasks and effectively addressed both conventional and complex challenges in the field.
LLMs are trained on multilingual~\cite{tang-etal-2024-language, wang2025multilinguallanguagemodelpretraining}, general-purpose~\cite{zhang-etal-2024-respond}, and web-based data, enabling them to generalize across languages~\cite{wu2025the}, handle interactions and code-switching~\cite{huzaifah-etal-2024-evaluating}, and flexibly respond to queries across a wide range of domains~\cite{wan-etal-2024-reformulating}.

Benchmarking initiatives are underway to evaluate LLM capabilities in language comprehension, generation, and reasoning, categorized by task types, domains, and languages~\cite{clark2018thinksolvedquestionanswering, wang2019gluemultitaskbenchmarkanalysis, park2021klue, zheng2023judging, hendrycks2021measuring}. These benchmarks critically enable objective comparisons among LLMs and identify areas for improvement. While existing benchmarks predominantly target general-purpose domains, such as MT-Bench~\cite{zheng2023judging}, MMLU~\cite{hendrycks2021measuring}, C-EVAL~\cite{ceval}, and Xiezhi~\cite{xiezhi}, specialized domains necessitate distinct problem-solving approaches and domain-specific knowledge, underscoring the increasing demand for benchmarks tailored to these fields.

General-domain benchmarks predominantly utilize standardized examination questions, which typically focus on STEM disciplines. Consequently, they offer limited insight into specialized knowledge domains and inadequately capture LLMs' domain-specific problem-solving capabilities. To address these gaps, recent benchmark studies have focused on detailed evaluations of LLM performance in specialized tasks requiring deep expert knowledge and practical application.

\begin{figure*}[!t]
    \centering
    \includegraphics[width=\linewidth]{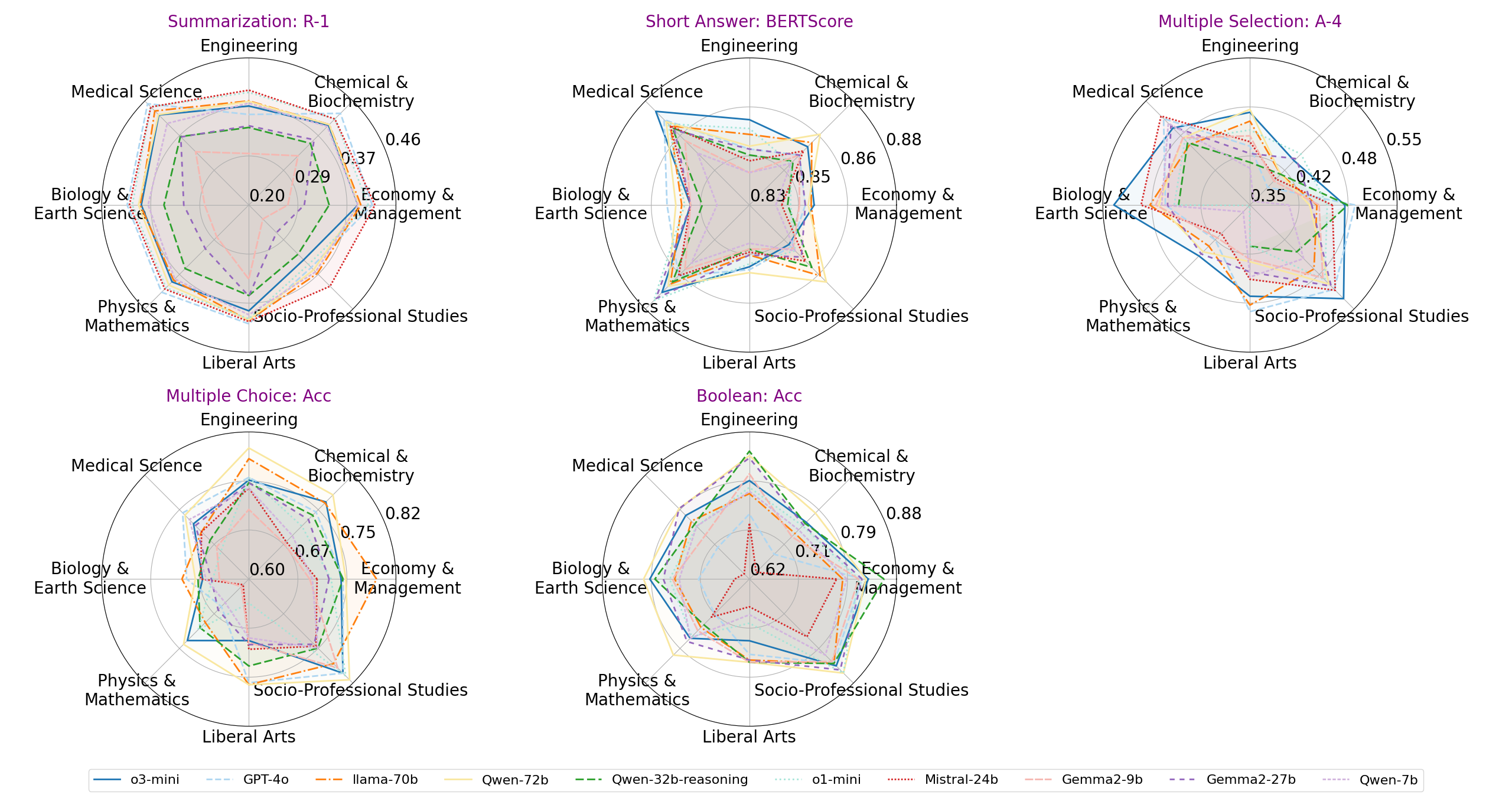}
    \caption{Model performance across categories for leading open- and closed-source LLMs on \texttt{ScholarBench}. Each column represents a task-specific evaluation metric. Main task-level results are reported in Table~\ref{tab:main-result}, while detailed performance analysis by category is provided in Appendix~\ref{appx:cate-anal}.}
    \label{fig:cate_eval}
\end{figure*}

For example, MultiMedQA~\cite{singhal2023large} and FinBen~\cite{xie2024finben} specifically assess LLM within medical and financial domains, demonstrating their practical applicability in tasks such as information extraction and risk management. Similarly, ChemLLMBench~\cite{guo2023what} and DataSciBench~\cite{zhang2024datascibench} provide comprehensive assessments within chemistry and data science domains, respectively, elucidating both strengths and limitations of current LLMs. Nonetheless, existing domain-specific benchmarks remain inherently constrained in their generalizability across disciplines, providing insufficient support for evaluating interdisciplinary and complex academic tasks.

In this paper, we introduce a new benchmark called \texttt{ScholarBench}, designed to evaluate the problem-solving capabilities of LLMs in academic domains, with a focus on their parametric knowledge, and analyze their performance in advanced reasoning tasks within scholarly environments. 
Academic-domain LLM benchmarks help enhance the practical applicability of LLMs in academic research, education, and specialized fields. \texttt{ScholarBench} offers three key features:

\setlength{\leftmargini}{1em} 
\begin{itemize}
    \vspace{-0.2cm}
    \item  
    \textbf{Domain and Attribute.}
    To systematically evaluate performance across interdisciplinary academic domains, we define four primary domains \textit{Natural Sciences}, \textit{Applied Sciences}, \textit{Social Sciences}, and \textit{Humanities} and further delineate a total of eight categories. 
    In addition, we propose fine-grained attribute categories to capture the diversity of question types and enable nuanced assessments of generalization performance within and across academic domains.     
    \vspace{-0.2cm}
    \item
    \textbf{Task and Evaluation.}
    \texttt{ScholarBench} leverages a diverse set of question types to concurrently assess multiple competencies of LLMs. Moving beyond simple item-level evaluation, the benchmark incorporates multidimensional assessment criteria including abstraction, reasoning, and comprehension to rigorously evaluate the practical problem-solving abilities and real-world applicability of LLMs. This approach offers a robust framework for holistic assessment of academic intelligence in LLMs.
    
    \vspace{-0.2cm}
    \item
    \textbf{Bilingual Ability.}
    We construct a bilingual (English–Korean) benchmark to enable precise evaluation of LLMs' cross-lingual knowledge transfer and multilingual understanding. 
    By designing questions and examples that facilitate both direct and indirect comparisons across linguistic and cultural contexts, we enable in-depth analysis of model generalization and performance disparities in the presence of linguistic diversity.
\end{itemize}

Overall, \texttt{ScholarBench} aims to assess LLMs' performance in academic domains, analyze question type-specific results from multiple perspectives to identify model strengths and weaknesses, and provide insights for improving LLMs in the academic domains.
The contributions of this paper are as follows:

\begin{figure*}[!t]
    \centering
    \includegraphics[width=\linewidth]{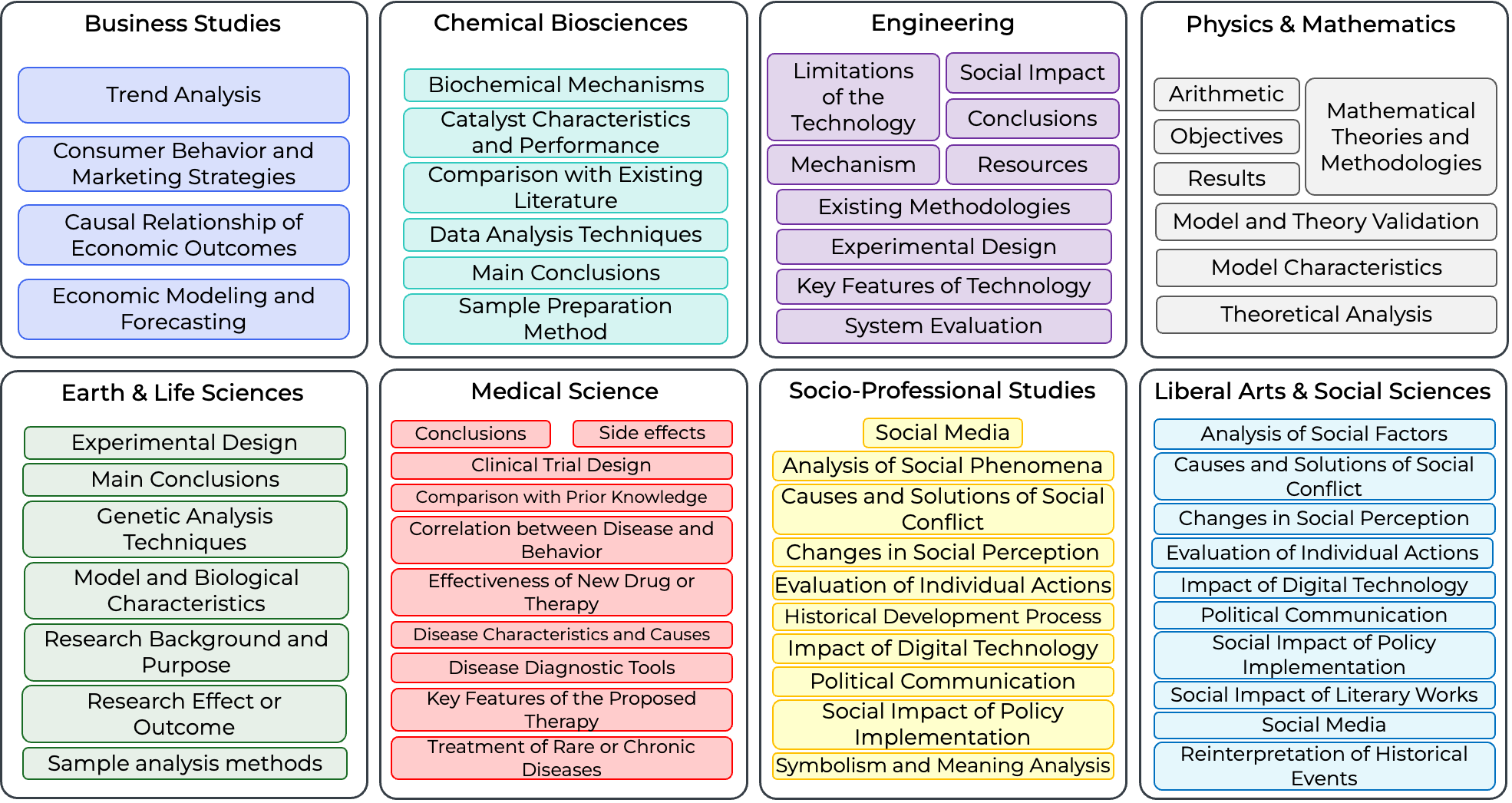}
    \caption{Taxanomy of academic categories and question attributes for English dataset.}
    \label{fig:categories}
\end{figure*}

\begin{itemize}
    \vspace{-0.2cm}
    \item Performance comparison of LLMs across eight academic categories, revealing domain-specific strengths and weaknesses
    \vspace{-0.2cm}
    \item Fine-grained evaluation of LLM capabilities across 63 English and 65 Korean academic attributes
    \vspace{-0.2cm}
    \item Construction of parallel and non-parallel bilingual datasets to assess linguistic and terminological understanding in academic texts
    \vspace{-0.2cm}
    \item A systematic benchmark construction method grounded in complex academic data
    \vspace{-0.2cm}
\end{itemize}
\section{Open Scholar Benchmark}
\noindent
We propose \texttt{ScholarBench}, a benchmark that enables performance evaluation and analysis of LLMs in academic domains, and introduce its construction methodology.
To define the multidimensional evaluation of LLMs, we present three key concepts:
\textbf{Abstraction (C1)} assesses the ability of LLMs to identify key information and summarize academic literature while maintaining the context.
\textbf{Reasoning ability (C2)} evaluates the LLM’s capacity for logical reasoning, demonstrated by its ability to infer answers based on parametric knowledge. The reasoning questions consist of short-answer, multiple-choice, multiple-selection, and true/false types, and are solved by the LLMs in a closed-book setting.
\textbf{Comprehension (C3)} is evaluated in an open-book setting, where the LLM must identify and extract key information from a given academic paragraph to determine its ability to solve academic problems accurately.

\subsection{Categories and Question Attributes} \label{sec:data_def}
\noindent
As shown in Figure~\ref{fig:categories}, we categorize eight academic categories and 63 English question attributes (along with 65 Korean attributes) into four academic domains:
\textit{natural sciences}, \textit{applied sciences}, \textit{social sciences}, and \textit{humanities}.
English papers are selected from journals with the highest H5-index according to Google Scholar~\footnote{https://scholar.google.com}, while Korean papers are selected from journals with the highest five-year citation index based on the KCI (Korea Citation Index)~\footnote{https://www.kci.go.kr/kciportal/main.kci}.
By integrating these sources, a total of eight categories are derived.
The attribute descriptions for each academic category are as follows:

\paragraph{Business Studies}
Focuses on analyzing key trends, consumer behavior, and marketing strategies in economic and business contexts. Tasks involve forecasting business or market performance through economic modeling and identifying causal relationships. Logical reasoning based on real-world data and case studies is essential.

\paragraph{Chemical Biosciences}
Covers the mechanisms of chemical and biochemical reactions, catalyst properties, interpretation of experimental results, and data analysis techniques. Emphasizes comparing findings with prior research and understanding sample preparation and analysis protocols. Scientific rigor and precise experimental design are critical.

\paragraph{Engineering}
Centers on engineering methodologies, technological innovation, experimental design, and system operations. Includes performance evaluation, resource efficiency, and social impact analysis to assess practical and sustainable use of technology. 

\paragraph{Physics \& Mathematics}
Emphasizes theoretical models, mathematical reasoning, and validation of physical systems. Tasks require evaluating model validity, achieving goals through experimental or computational results, and solving quantitative problems through logical deduction.

\paragraph{Earth \& Life Sciences}
Addresses topics in life sciences, including biological modeling, genetic analysis, and experimental procedures. Combines theoretical understanding and empirical methods to analyze causality and correlations. 
Combines theoretical biological modeling with analyses to elucidate quantitative relationships in biological phenomena.

\paragraph{Medical Science}
Deals with practical issues in healthcare, including clinical trial design, diagnostic tools, drug efficacy, and side effect analysis. Encompasses rare and chronic disease treatments and preventive strategies. Tasks require integrating scientific evidence with medical reasoning.

\begin{figure*}[!t]
    \centering
    \includegraphics[width=\linewidth]{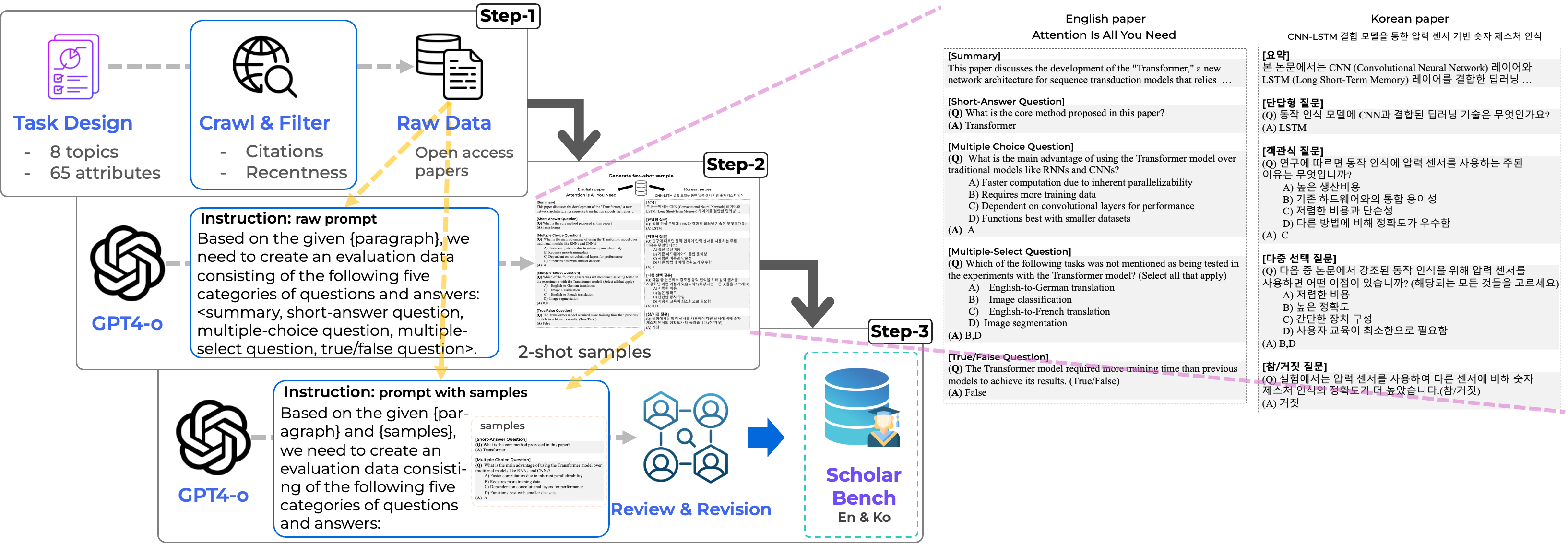}
    \caption{
    Data construction pipeline. For a step-by-step example of data construction, see Appendix~\ref{appx:data_construction}.
    }
    \label{fig:cons_pipeline}
\end{figure*}

\paragraph{Socio-Professional Studies}
Covers topics at the intersection of society and professional practice, such as the effects of arts and physical education, psychological factors, and interpretation of cultural artifacts.

\paragraph{Liberal Arts \& Social Sciences}
Explores causes and transformations of social phenomena, policy impact, and the role of digital media. Includes humanities disciplines such as literature, history, politics, and philosophy. Tasks emphasize critical thinking, interpretation, and understanding individual and collective behavior in social contexts.

\subsection{Question Design}
\noindent
Academic-domain benchmarks aim to evaluate whether LLMs can apply their learned knowledge to reason about and solve domain-specific problems.
The proposed benchmark includes five types of questions: summarization, short answer, multiple choice, multiple selection, and true/false. These question types are designed not only to assess LLMs’ abstraction, reasoning, and comprehension abilities, but also to ensure diversity, generality, and ease of evaluation.

To evaluate whether LLMs can adequately perform academic-domain tasks using only their pretrained knowledge, we conduct experiments under both closed-book and open-book settings. In the closed-book setting, no external information is provided except for summarization tasks requiring the model to rely solely on its internal knowledge. To ensure evaluation diversity and analyze the impact of information access on performance, we additionally include the open-book setting.

Among the five question types, summarization tasks require the model to condense a given paragraph into its key content. 
Short-answer questions involve understanding the query and generating a specific, correct response~\cite{rajpurkar-etal-2018-know}.
Multiple-choice questions assess the model’s decision-making based on provided options.
Multiple-selection questions demand higher discriminative ability, as the model must identify all correct answers.
Finally, true/false questions, commonly used in benchmarks such as BoolQ~\cite{clark2019boolq} and TruthfulQA~\cite{lin-etal-2022-truthfulqa}, evaluate factual reasoning by requiring binary judgments based on a given question.

These question types are comparable to those used in general-purpose benchmarks, allowing us to assess LLMs' problem-solving capabilities in specialized domains and to estimate the difficulty of the benchmark.

\subsection{Data Construction Process}
\noindent
To ensure high-quality question-answer pairs aligned with each academic domain, we conduct a three-step data construction pipeline.
Figure~\ref{fig:cons_pipeline} illustrates the overall pipeline, which consists of the following steps.
In step 1, tasks are designed based on eight academic categories and 63 English and 65 Korean attributes, and source materials are collected by crawling and filtering publicly available academic papers according to recency and citation count.
In step 2, five types of evaluation questions, summarization, short-answer, multiple-choice, multiple-selection, and true/false, are generated from the source paragraphs using GPT-4o. 
In step 3, to ensure varying question types and difficulties across turns, two questions from the initial outputs are randomly sampled to iteratively refine the prompting strategy. 
Finally, the generated questions undergo expert review and revision to improve quality before being finalized in \texttt{ScholarBench}. 
Motivated by our observation that question diversity and difficulty vary depending on the source documents, we performed a comprehensive evaluation integrating passages, generated questions, and corresponding answers.
Detailed prompts, data samples, and diversity-related statistics used for question generation are available in Appendix~\ref{appx:prompt-template}.

\paragraph{Academic Papers Collection.}
The criteria for selecting academic papers are as follows.
For English papers, we collect a total of 200 articles from 80 Open Access journals that fall under our academic categories and are ranked in the top 10 of the H5-index according to Google Scholar.
For Korean papers, we collect 1,916 articles from 138 journals selected using the same criteria, based on the KCI.

\paragraph{Review and Revision.}
The data verification process is conducted by 15 reviewers, 11 for English and 4 for Korean. All reviewers are university students, and they follow the guidelines.
The review process consists of three stages:
\textbf{Paragraph evaluation}, where reviewers assess whether the selected paragraph appropriately reflects the core concept of the paper;
\textbf{Answer verification}, where the correctness of the provided answer for each item is checked;
\textbf{Human evaluation}, which is conducted only if the answer is correct, to assess the overall quality of the example.
The main focus points during the review process are as follows:

\setlength{\leftmargini}{1em} 
\begin{itemize}
    \vspace{-0.2cm}
    \item  
    All five question types are derived from a single paragraph, and the evidence required to answer each question must be explicitly present in the paragraph.
    \vspace{-0.2cm}
    \item
    Questions should be designed to avoid the use of referential noun phrases (e.g., demonstratives or determiners) that do not directly point to specific content in the paragraph. Due to certain attributes such as research objectives and conclusions (see Table~\ref{fig:categories}), expressions like “this study” may be automatically generated. During the review process, we ensure such referential phrases are excluded from the question prompt.    
    \vspace{-0.2cm}
    \item
    In \texttt{ScholarBench}, short-answer questions are designed to have a single correct answer. Ambiguous cases that could allow multiple valid answers are excluded, enabling evaluation of whether the LLM can generate a precise response. This design is important for assessing reasoning ability with parametric knowledge.
\end{itemize}

\begin{table}[]
\centering
\small
\resizebox{0.48\textwidth}{!}{
    \begin{tabular}{l|ccc|c|c}
    \toprule
    \multirow{2}{*}{\begin{tabular}[c|]{@{}l@{}}Language\end{tabular}} & \multicolumn{3}{c|}{Assessment} & \multirow{2}{*}{Evaluation}  & \multirow{2}{*}{\begin{tabular}[c]{@{}l@{}}Kappa Coefficient\end{tabular}} \\ 
             & A & B & C & \\
    \midrule
    English  & 4.56 & 4.47 & 4.36 & 0.72 & 0.621 \\
    Korean  & 4.6 & 4.73 & 4.67 & 0.7 &  0.742 \\
    \bottomrule
    \end{tabular}
}
\caption{Human evaluation and data difficulty assessment using a 1–5 scale and Kappa coefficient.}
\label{tab:data-human-eval}
\end{table}

To measure inter-annotator agreement during the benchmark validation process, we report the evaluation scores given by three annotators across two evaluation dimensions—\textit{Assessment} and \textit{Evaluation}—as shown in Table~\ref{tab:data-human-eval}. 
For this, we randomly sample 9.1\% of the examples in each language, which were independently assessed by the annotators, with overlapping cases cross-annotated by three annotators.
The resulting Kappa coefficients~\cite{kappa} indicate a high level of agreement: 0.621 for English and 0.742 for Korean, demonstrating that \texttt{ScholarBench} provides consistent and reliable annotations.

\section{Experiments}
\noindent
In this chapter, we evaluate the performance of publicly available English-Korean bilingual models to determine the validity of the \texttt{ScholarBench}.

\subsection{Target Models}
\noindent
\begin{itemize}
    \item \textbf{Three API-based models} include o3-mini, o1-mini, GPT-4o~\cite{openai2024gpt4ocard}
    \item \textbf{Six open-source model families} include Llama3.3~\cite{grattafiori2024llama3herdmodels}, Mistral~\cite{jiang2024mixtralexperts}, Qwen2.5~\cite{qwen2.5}, Gemma2~\cite{gemmateam2024gemma2improvingopen}, Bllossom,  Exaone~\cite{research2024exaone35serieslarge}.
\end{itemize}
A detailed description of each model can be found in the Appendix~\ref{appx:model-card}.

\subsection{Evaluation Metrics}
\noindent
We use ROUGE~\citep{lin-2004-rouge} to evaluate summarization tasks, and accuracy to assess performance on multiple-choice, multiple-selection, and boolean questions.
For short-answer questions, we use BERTScore\footnote{We use the 'xlm-roberta-base' checkpoint\footnote{\url{https://huggingface.co/FacebookAI/xlm-roberta-base}} provided by Facebook AI.}~\citep{zhang2020BERTScore} to assess semantic similarity. 
We measure accuracy in multiple-selection, multiple-choice, and boolean questions. 
In multiple selection evaluation, performance is assessed by varying the target number of correct answers (2-4). For a target of 2, an accuracy point is given upon selecting any two correct options, irrespective of the total options presented (e.g., 2 out of 4 total).
The evaluation metrics used in this paper are shown in Appendix~\ref{appx:metrics}.

\subsection{Overall Performance}
\begin{table*}[!t]
    \centering
    \tiny
    \resizebox{\textwidth}{!}{
    \begin{tabular}{lcccccccccc}
    \toprule
    \multirow{2}{*}{\begin{tabular}[c]{@{}l@{}}Model\end{tabular}} & \multicolumn{3}{c}{Summarization}		& {Short Answer}            & \multicolumn{3}{c}{Multiple Selection}	& \multirow{2}{*}{\begin{tabular}[c]{@{}l@{}}MCQ\end{tabular}} & \multirow{2}{*}{\begin{tabular}[c]{@{}l@{}}Boolean\end{tabular}} & \multirow{2}{*}{\begin{tabular}[c]{@{}l@{}}Avg\end{tabular}} \\ \cmidrule(lr){2-4} \cmidrule(lr){5-5} \cmidrule(lr){6-8} 
      & R-1	& R-2	& R-L	 & BERTScore	& A-2	& A-3	& A-4	&    &         &     \\ \hline
    o3-mini  & 0.392 & 0.130 & 0.320 & \bf 0.860 & \bf 0.666 & 0.518 & \bf 0.482 & 0.728 & 0.786 & \bf 0.543 \\ 
    o1-mini & \underline{0.409} & \underline{0.151} & 0.348 & 0.857 & 0.567 & 0.469 & 0.436 & 0.702 & 0.771 & 0.523 \\ 
    GPT-4o  & \bf 0.417 & \underline{0.151} & \underline{0.356} & 0.857 & 0.586 & 0.497 & \underline{0.465} & 0.736 & 0.743 & \underline{0.534} \\ 
    Qwen-72b & 0.396 & \underline{0.151} & 0.345 & \underline{0.859} & 0.522 & 0.468 & 0.452 & \bf 0.755 & \bf 0.811 & 0.529 \\ 
    Llama-70b & 0.397 & 0.146 & 0.341 & 0.858 & 0.578 & 0.485 & 0.455 & \underline{0.746} & 0.769 & 0.530 \\ 
    Bllossom-70b & 0.349 & 0.129 & 0.299 & 0.848 & \underline{0.650} & 0.463 & 0.440 & 0.683 & 0.724 & 0.509 \\ 
    Qwen-32b-reasoning & 0.350 & 0.116 & 0.303 & 0.853 & 0.539 & \textbf{0.640} & 0.423 & 0.721 & 0.793 & 0.527 \\ 
    Exaone-32b & 0.321 & 0.094 & 0.267 & 0.852 & 0.590 & 0.466 & 0.431 & 0.713 & 0.751 & 0.498 \\ 
    Exaone-32b-reasoning & 0.316 & 0.092 & 0.267 & 0.847 & 0.492 & 0.397 & 0.359 & 0.686 & 0.667 & 0.458 \\ 
    Gemma2-27b & 0.329 & 0.117 & 0.283 & 0.856 & 0.577 & 0.479 & 0.453 & 0.707 & \underline{0.796} & 0.511 \\ 
    Mistral-24b & 0.414 & \bf 0.159 & \bf 0.359 & 0.853 & 0.584 & 0.488 & 0.458 & 0.694 & 0.696 & 0.523 \\ 
    Gemma2-9b & 0.294 & 0.096 & 0.248 & 0.851 & 0.556 & \underline{0.520} & 0.444 & 0.684 & 0.774 & 0.496 \\ 
    Exaone-8b & 0.317 & 0.092 & 0.265 & 0.846 & 0.577 & 0.417 & 0.386 & 0.692 & 0.756 & 0.483 \\ 
    Mistral-8b & 0.402 & \underline{0.151} & 0.350 & 0.842 & 0.504 & 0.374 & 0.355 & 0.656 & 0.582 & 0.468 \\ 
    Llama-8b & 0.381 & 0.136 & 0.327 & 0.845 & 0.501 & 0.419 & 0.395 & 0.658 & 0.556 & 0.469 \\ 
    Bllossom-8b & 0.346 & 0.129 & 0.301 & 0.844 & 0.537 & 0.419 & 0.383 & 0.633 & 0.581 & 0.464 \\ 
    Qwen-7b & 0.388 & 0.144 & 0.338 & 0.847 & 0.559 & 0.452 & 0.423 & 0.699 & 0.756 & 0.512 \\ 
    \bottomrule 
    \end{tabular}
    }
    \caption{Overall evaluation results of \texttt{ScholarBench} on all curated prompts under closed-book settings. For performance metrics, \textit{R} denotes ROUGE scores (ROUGE-1, ROUGE-2, and ROUGE-L), and \textit{A} indicates multiple selection settings, where the appended number signifies the count of correct answers. Only the summarization task is evaluated using paragraph-level input; all subsequent tasks are evaluated without paragraph-level input. The \textit{Avg} column reports the average over all listed metrics.
    Bold and underline indicate the first and second ranks per metric.}
    \label{tab:main-result}
\end{table*}
\noindent
Table~\ref{tab:main-result} presents the evaluation results for API-based and open-source models on \texttt{ScholarBench}. 
The evaluation comprises five tasks: summarization, short-answer, multiple-selection, multiple-choice question (MCQ), and boolean question, showing representative metrics for each task. 
This setup allows for identifying performance variations across tasks and specialization tendencies of models. 
The table includes the average across all metric results to provide an overview of the models' overall performance. 

\paragraph{Summarization.}
In summarization tasks, GPT-4o and Mistral-24b demonstrate strong overall performance.
Specifically, GPT-4o achieves the top score in ROUGE-1, while Mistral-24b leads in both ROUGE-2 and ROUGE-L.
We hypothesize that while GPT-4o is effective in capturing simple information, Mistral-24b excels in comprehensively processing longer context.
Furthermore, among the small LLMs, Mistral-8b exhibits performance comparable to o1-mini, suggesting that even a small model can achieve competitive results depending on the architecture design and training strategy.

\paragraph{Short Answer.}
The short-answer task reveals only marginal performance differences among top models, and most of them show a high BERTScore of 0.84 or higher.
This appears to be because semantic similarity-based evaluation is more sensitive to the naturalness and consistency of expressions than to complex inferences.
In addition, small LLMs show relatively score in this task, suggesting that the models can be adjusted to trade-off balance between model efficiency and accuracy.

\paragraph{Multiple Selection.}
This task exhibits a tendency for model performance to vary as the number of options increases.
Specifically, o3-mini achieves the best performance in the A-2 setting, while Qwen-32b-reasoning excels in A-3, indicating that the optimal model differs depending on the number of options.
A general performance degradation is observed in the A-4 setting, implying that as the number of correct answers increases, the difficulty in achieving partial correctness or the challenge of avoiding distractors increases.
The strong performance of Qwen-32b-reasoning in A-3 suggests that reasoning-specific tuning can be effective for addressing specific types of complex multiple selection questions.

\paragraph{MCQ.}
In the MCQ, Qwen-72b, Llama-70b, GPT-4o achieve top scores. 
These models increase accuracy in answer selection, due to their ability to effectively discriminate between candidate options based on semantic similarity and identify distractor.
On the other hand, models under 8b parameters exhibit decreased performance on this task.
These results reflect a correlation between model size and problem-solving ability in the MCQ task, which requires logical comparitive judgement, reasoning, long context dependecy.

\paragraph{Boolean.}
On this task, Qwen-72b demonstrates top performance, achieving a score of 0.811, which highlights its strong capability for binary classification.
Qwen-32b-reasoning and Gemma2-27b also achieve top-tier accuracy, there is a significant contribution of explicit reasoning-oriented fine-tuning to performance on this task.
The significantly decreased accuracy of small LLMs suggests that model parameter size considerably impacts performance.
However, among small LLMs, Gemma2-9b, Exaone-8b, and Qwen-7b maintain scores above 0.75, indicating that problem-solving ability can be improved through learning strategy even with a small parameter size.

\paragraph{Average Performance.}
The average performance across all metrics (the \textit{Avg} column) reveals well-balanced capabilities encompassing generation, understanding, classification, and reasoning.
As shown in Table~\ref{tab:main-result}, o3-mini achieves the highest average score (0.543), followed by models such as GPT-4o, o1-mini, Qwen-72b, Llama-70b, Qwen-32b-reasoning, Mistral-24b, and Gemma2-9b, which also demonstrate top-tier performance with scores above 0.52.
LLMs exhibit stable and consistent performance, even on academic domain-specific tasks.
Meanwhile, some medium and small-sized models also show significant average scores. 
Accordingly, this highlights that \texttt{ScholarBench} was designed not for a biased comparison centered on a single task, but to evaluate various problem types and difficulty levels.

\section{Analysis}
\begin{table}[t]
\centering
\small
\setlength{\tabcolsep}{6pt}
\begin{tabular}{lcccc}
\toprule
Model & R-1 & R-2 & R-L & BERTScore \\
\hline
\multicolumn{5}{c}{\textit{Paragraph-Only Prompting}} \\ \hline
o1-mini & 0.409 & 0.151 & 0.348 & 0.885 \\
Qwen-72b & 0.396 &	0.151 &	0.345 &	0.879 \\
Mistral-8b & 0.402	& 0.151	& 0.350 &	0.883 \\
\hline
\multicolumn{5}{c}{\textit{Prompting with Paragraph and Category}} \\ \hline
o1-mini & 0.413 & 0.152 & 0.349 & 0.886 \\
Qwen-72b & 0.412 &	0.160 &	0.357 &	0.882 \\
Mistral-8b & 0.408 &	0.155 &	0.355 &	0.884 \\
\bottomrule
\end{tabular}
\caption{Evaluating abstraction ability using summarization under different prompting settings.}
\label{tab:abs_ability}
\end{table}

\begin{table*}[t]
\centering
\small
\resizebox{\textwidth}{!}{
\begin{tabular}{p{0.27\linewidth} p{0.32\linewidth} p{0.2\linewidth} p{0.13\linewidth}}
\toprule
\textbf{Passage (excerpt)} & \textbf{Question} & \textbf{Answer} & \textbf{Reasoning} \\
\midrule
\multirow{4}{=}{
\textit{Dominican youth of Haitian descent face significant barriers to education due to lack of documentation, societal prejudice, and economic hardship. Many children receive Dominican birth certificates, only to have them cancelled arbitrarily. High dropout rates are observed, especially for females, who may encounter additional challenges from traditional gender roles. Despite these adversities, some students continue their education.}
} 
& What specific challenge related to documentation impacts school participation for Dominican females of Haitian descent? & Lack of documentation & \textbf{Yes} \\ \cmidrule{2-4}
& What factors contribute to the educational challenges faced by Dominican girls of Haitian descent? (Select all that apply) & a) Cultural attitudes like machismo, b) Economic hardship & \textbf{Yes} \\ \cmidrule{2-4}
& What impact can the lack of documentation have on youth education? & a) Denial of access to national exams & \textbf{Yes} \\ \cmidrule{2-4}
& The absence of documentation does not affect the educational success of Dominican females of Haitian descent. (True/False) & False & \textbf{Yes} \\
\bottomrule
\end{tabular}
}
\caption{Example of a reasoning-based question set. The passage is excerpted to retain core information about cultural capital, systemic oppression, and education resilience, allowing evaluation of LLMs' contextual reasoning.}
\label{tab:reasoning-example}
\end{table*}

\begin{table*}[!t]
    \centering
    \tiny
    \resizebox{\textwidth}{!}{
    \begin{tabular}{lcccccccccc}
    \toprule
    \multirow{2}{*}{\begin{tabular}[c]{@{}l@{}}Model\end{tabular}} & \multicolumn{3}{c}{Summarization} & Short Answer     & \multicolumn{3}{c}{Multiple Selection}	& \multirow{2}{*}{\begin{tabular}[c]{@{}l@{}}MCQ\end{tabular}} & \multirow{2}{*}{\begin{tabular}[c]{@{}l@{}}Boolean\end{tabular}}	   & \multirow{2}{*}{\begin{tabular}[c]{@{}l@{}}Avg\end{tabular}}	   \\ \cmidrule(lr){2-4} \cmidrule(lr){5-5} \cmidrule(lr){6-8}
         & R-1 & R-2 & R-L	 & BERTScore	& A-2	& A-3	& A-4	&    &	&	   \\ \hline
        o3-mini  & 0.392 & 0.130 & 0.320 & 0.898 & 0.783 & 0.611 & 0.591 & 0.886 & 0.875 & 0.609 \\
        o1-mini & 0.409 & 0.151 & 0.348 & 0.898 & \textbf{0.795} & 0.656 & 0.632 & 0.883 & 0.872 & 0.627 \\
         GPT-4o  & \textbf{0.417} & \textbf{0.151} & 0.356 & 0.897 & 0.793 & \textbf{0.667} & \textbf{0.643} & 0.886 & 0.854 & \textbf{0.629} \\
        Qwen-72b & 0.396 & 0.151 & 0.345 & 0.902 & 0.746 & 0.650 & 0.628 & \textbf{0.900} & \textbf{0.896} & 0.624 \\
        Llama-70b & 0.397 & 0.146 & 0.341 & \textbf{0.903} & 0.751 & 0.620 & 0.598 & 0.875 & 0.873 & 0.612 \\
        Mistral-24b & 0.414 & 0.159 & \textbf{0.359} & 0.901 & 0.756 & 0.580 & 0.560 & 0.865 & 0.827 & 0.602 \\
     \bottomrule
    \end{tabular}
    }
    \caption{Evaluation of comprehension ability in \texttt{ScholarBench} under open-book settings: effect of paragraph prompting on task performance. Summarization results correspond to ROUGE-1 scores as reported in Table~\ref{tab:main-result}.}
    \label{tab:comprehension_eval}
\end{table*}

\noindent
In this section, we evaluate and analyze the abstraction, comprehension, reasoning, and bilingual abilities. For more detailed experimental results, please refer to Appendix~\ref{appx:full-eval}.

\subsection{LLM Capability Analysis}
\noindent
\paragraph{Abstraction}
For abstraction (summarization) tasks, we compare the performance variations in selected models depending on whether a category was added to the prompt.
In Table~\ref{tab:abs_ability}, the models demonstrate good summarization performance even on academic domain passages.
Furthermore, when prompting with a domain-specific category, a slight but noticeable performance gain is observed.

\paragraph{Comprehensibility}
\texttt{ScholarBench} includes paragraphs supporting each question for the evaluation of comprehension ability. 
This feature is aimed at measuring the ability to extract key information from the provided passage and generate correct responses. 
Table~\ref{tab:comprehension_eval} presents the performance results when prompting with both the question and the paragraph. 
Overall performance improves for all models, indicating that the inclusion of paragraphs contributes to improving the models' problem-solving capabilities, and suggesting that the models are able to understand the paragraph well and infer the correct answer. 
For LLMs, not only simple reasoning ability but also the ability to interpret and utilize the given context is crucial. Thus, a benchmark designed to evaluate comprehension ability is valuable.

\paragraph{Reasoning}
Table~\ref{tab:reasoning-example} presents an example of a reasoning question, answer, and the supporting paragraph.
Based solely on simple information explicitly listed in the paragraph, deriving the correct answer to the question \textit{What factors contribute to the educational challenges faced by Dominican girls of Haitian descent?} is difficult.
Accurate selection for this question requires contextual inference and integration of the impact of factors such as economic difficulties and gender roles and cultural norms on education.
This question type evaluates models' ability to relationally connect detailed information within the paragraph and their understanding of the background and context.
Additionally, \textit{What impact can the lack of documentation have on youth education?} requires causal inference between documentation issues and education.
Although this information is not explicitly written in the paragraph, it necessitates contextual reasoning such as \textit{lack of documentation} → \textit{inability to take exams} → \textit{exclusion from education}.
The boolean question presents the core argument of the paragraph in reverse, and measures judgment for logical negation.

~\begin{figure}[!t]
    \centering
    \includegraphics[width=\linewidth]{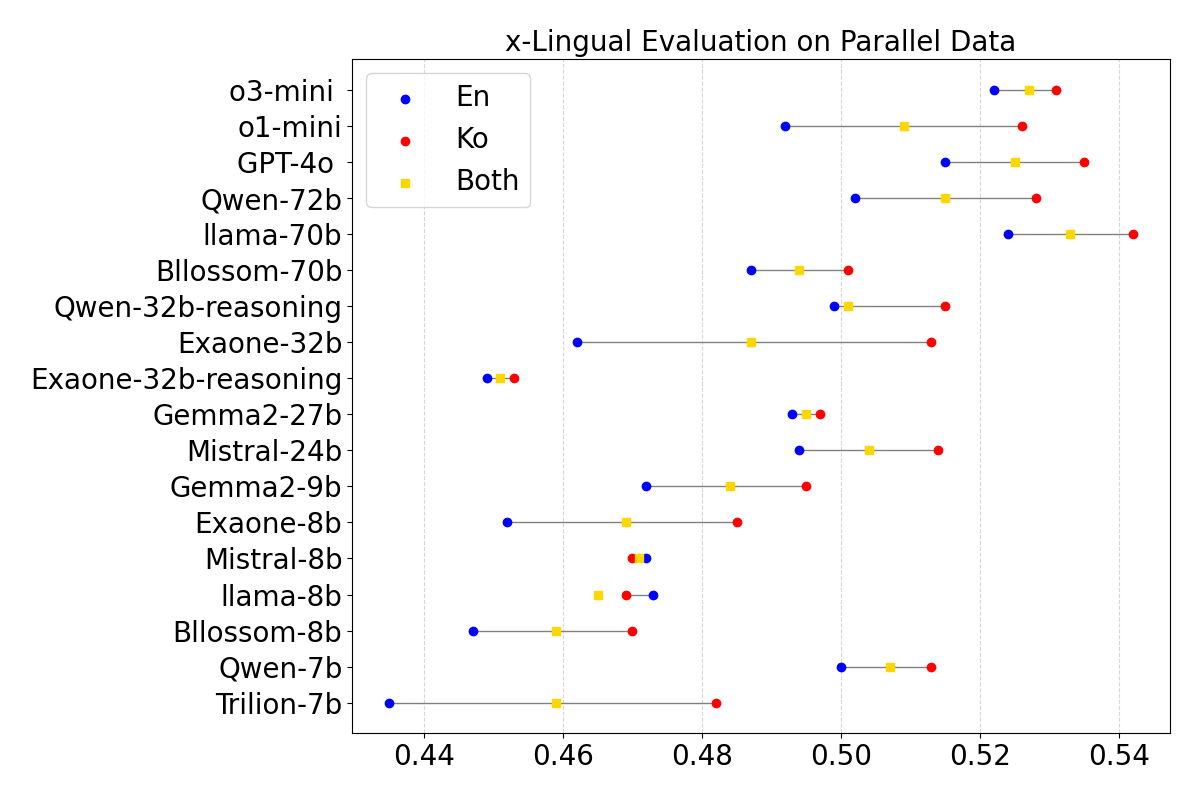}
    \caption{Model-wise performance on parallel data across En, Ko, and Both language settings.}
    \label{fig:parallel_eval}
\end{figure}
\paragraph{Bilingual Ability}
\texttt{ScholarBench} is partially constructed as parallel data in English and Korean for 18.7\% of all questions.
Figure~\ref{fig:parallel_eval} shows the results of cross-lingual evaluation based on this parallel data. 
This experiment analyzes the cross-lingual generalization capabilities of multilingual LLMs, reflecting complex factors such as semantic consistency across languages, reasoning coherence, and potential information loss during translation. 
Most models exhibit a Ko > En trend, while some global models show the opposite pattern (En > Ko).
\section{Conclusion}
\noindent
We propose \texttt{ScholarBench}, a new benchmark dataset for evaluating the capabilities of LLMs in academic domains, and analyze the experimental results.
While previous benchmark's studies focus on specific tasks or general domains, \texttt{ScholarBench} is designed to include various abilities and question types required in academic fields, such as abstraction, comprehension, reasoning, and bilingual ability.
By defining eight academic domain categories and 63 corresponding attributes, we enhance the diversity of problems and provide a guide for more detailed and systematic evaluation of problem-solving abilities.

As analysis results utilizing  \texttt{ScholarBench}, we confirmed the significant variations that LLMs show depending on the task types, academic fields, and model characteristics.
For each task, we analyze model performance based on the problem characteristics such as reliance on structured information, context dependency, expression diversity, causal reasoning, relational understanding, and logical negation. 
This enables the evaluation of both \textit{topic generalization ability} and \textit{topic-specific reasoning ability} to handle the complex knowledge and reasoning within specific academic fields.

For future work, we envision further expanding \texttt{ScholarBench} to include more academic domains and new task types, such as RAG and multimodal.
Additionally, we plan to specifically analyze model error types to identify the root causes of performance degradation and utilize \texttt{ScholarBench} to develop LLM learning and tuning strategies specialized for academic domains.
Finally, we expect that \texttt{ScholarBench} will contribute to the development of more capable and reliable AI systems for the academic community.

\section*{Limitations}
While \texttt{ScholarBench} provides a comprehensive evaluation of academic problem-solving across various fields, its scope is largely confined to English-language academic settings and structured tasks. 
This means it may not fully capture the reasoning challenges unique to multilingual or less formal educational contexts, thus limiting its generalizability beyond standardized academic domains.

Our constructed benchmark dataset, \texttt{ScholarBench}, was designed to evaluate academic problem-solving abilities based on scholarly articles. However, a primary limitation is that the data sources were confined solely to paragraph text from these articles. Real scholarly papers, beyond plain text, contain information across diverse modalities such as figures, tables, algorithms, and diagrams, which are often essential for understanding the research content. Evaluating models on academic paper data typically requires the ability to integrate and reason over information from various modalities. Since the current dataset does not explicitly incorporate or utilize these multimodal elements, this poses a direct limitation in accurately measuring the academic problem-solving capabilities of models that leverage multimodal information. Future work is needed to overcome this limitation by constructing an extended dataset that includes and utilizes diverse modalities.

To effectively evaluate the model's comprehensibility of specific text passages, \texttt{ScholarBench} is structured such that paragraphs related to each question are provided as prompts alongside the question. This setup is valuable for assessing how accurately a model interprets and responds based on given contextual information. However, a recent major trend in NLP, such as retrieval-augmented generation (RAG)~\cite{rag-origin, gao2024retrievalaugmentedgenerationlargelanguage, li-etal-2024-retrieval}, emphasizes the overall pipeline performance, which includes finding relevant external knowledge and generating responses based on it. The current dataset structure presents a limitation in that it cannot directly evaluate the performance of the Retrieval phase necessary before problem-solving in a RAG setup. As future work, we plan to leverage the currently developed questions but extend the evaluation process to require models to first retrieve and extract the necessary information from the original full papers or a collection of related documents.

\section*{Acknowledgements}
This research was supported by the Korea Institute of Science and Technology Information (KISTI) in 2024 (No.(KISTI) K24L4M2C5), (NTIS) 2710007168), aimed at developing KONI (KISTI Open Neural Intelligence), a large language model specialized in science and technology.
This work was supported by Institute of Information \& communications Technology Planning \& Evaluation (IITP) grant, funded by the Korea government (MSIT) (No.RS-2024-00456709)
This research was supported by the 2025 Daejeon RISE Project, funded by the Ministry of Education and Daejeon Metropolitan City, in the Republic of Korea.

\section*{Ethical Considerations}
Resource is available at \url{https://huggingface.co/datasets/KISTI-KONI/ScholarBench}

\paragraph{Copyright and License.}
The \texttt{ScholarBench} will be distributed under the CC BY-ND 4.0 license.
\bibliography{custom}

\newpage
\appendix

\section{Related Work}
\noindent
Benchmark studies have focused on evaluating specific capabilities or domains, including domain-specific benchmarks, multilingual and cultural benchmarks, and multi-domain knowledge benchmarks.

\paragraph{Domain-Specific Benchmarks}
Domain-specific benchmarks aim to evaluate whether large language models (LLMs) can go beyond general language understanding to comprehend and apply specialized knowledge and terminology within specific fields.
In the scientific domain, benchmarks such as SciNLI~\cite{sadat2022scinlicorpusnaturallanguage}, MSciNLI~\cite{sadat2024mscinlidiversebenchmarkscientific}, and Matsci-NLP~\cite{song2023matscinlpevaluatingscientificlanguage} have been introduced.
In the medical domain, MedMCQA~\cite{pal2022medmcqalargescalemultisubject} has been widely adopted, and FinBen~\cite{xie2024finben} is proposed for the financial domain.
Unlike general-domain NLI datasets, SciNLI evaluates scientific language understanding based on research papers. 
Compared to other NLI datasets, a characteristic feature is the low vocabulary overlap between the premise and hypothesis, which necessitates a deeper understanding rather than reliance on surface-level lexical cues.
MSciNLI extends SciNLI by covering domains such as hardware, networks, software engineering, and security, thus increasing data diversity.
Matsci-NLP focuses on evaluating LLM performance in the materials science domain. The dataset consists of seven tasks covering topics such as superconductors, fuel cells, and glass.
MedMCQA is a large-scale multiple-choice QA benchmark constructed from actual medical entrance exams.
MedMCQA covers 2,400 medical topics and 21 medical subjects, and contains a total of more than 194,000 entrance exam questions. Each data point includes a question, correct and incorrect answer options, and evaluates 10 or more reasoning skills to assess language comprehension skills in medical subjects and topics.
FinBen is designed to assess the capabilities of LLMs in financial tasks. This dataset contains 36 different data sets spanning 24 financial tasks, including information extraction, text analysis, forecasting, and risk management.

\paragraph{Multilingual and Culturally-Aware Benchmarks}
With the growing adoption of LLMs by diverse user groups, understanding cultural context and supporting low-resource languages have become increasingly important.
ALM-Bench~\cite{vayani2024languagesmatterevaluatinglmms} evaluates LLMs’ ability to understand and reason over text and culturally grounded images across 100 languages. 
This benchmark is characterized by its ability to comprehensively evaluate cultural characteristics, including culturally diverse features ranging from low-resource languages to specific regional dialects.
CVQA~\cite{romero2024cvqaculturallydiversemultilingualvisual} presents a culturally diverse visual question answering benchmark consisting of questions and images collected from 30 countries across four continents. Each question is written in both English and the local language, allowing assessment of both multilingual and English-only models.

\paragraph{Multi-Domain Knowledge Benchmarks}
Multi-domain knowledge benchmarks evaluate the generalization capabilities of LLMs by testing them across a wide range of subjects.
These benchmarks aim to assess how well LLMs perform in diverse academic and professional areas.
Notable datasets include MMLU~\cite{hendrycks2021measuring}, Xiezhi~\cite{xiezhi}, and AGIEval~\cite{zhong2023agievalhumancentricbenchmarkevaluating}.
MMLU includes problems on diverse topics across 57 domains and is composed of multiple-choice questions of varying difficulty. 
It evaluates the comprehensive understanding and problem-solving abilities of large language models by assessing how well they understand diverse knowledge. 
Xiezhi includes 13 subjects and 516 diverse fields, and consists of a total of 249,587 multiple-choice questions. 
To evaluate multiple-choice questions, it uses Mean Reciprocal Rank (MRR)~\cite{sirotkin2013searchengineevaluationmetrics}, which calculates a ranking score, as the evaluation metric. 
A notable feature is the continuous updating of the benchmark by automatically generating data from open academic resources and labeling it with trained models. 
AGIEval is a benchmark dataset that evaluates the performance of large language models based on human-centric standardized tests such as college entrance exams, law school admission tests, mathematics competitions, and other professional qualification exams. It excludes subjective questions and includes objective questions like multiple-choice and fill-in-the-blank.

\paragraph{Evaluating Conversational Capabilities}
Existing datasets for evaluating the conversational capabilities of large language models include MT-Bench~\cite{zheng2023judging}, MT-Eval~\cite{kwan2024mtevalmultiturncapabilitiesevaluation}, and CLIcK~\cite{CLIcK}.
MT-Bench consists of 80 high-quality multi-turn questions designed to evaluate conversational flow and instruction following. 
It covers 8 task categories: writing, role-playing, extraction, reasoning, math, coding, knowledge I (STEM), and knowledge II (humanities/social science). 
Each task category comprises 10 multi-turn problems. 
Model outputs are obtained for these tasks, and the model responses are then evaluated based on criteria including context understanding, accuracy, consistency of reasoning steps, and whether responses meet user expectations. 
MT-Eval is an extended benchmark of MT-Bench that refines the evaluation of multi-turn conversational capabilities by evaluating abilities such as remembering and utilizing previously mentioned information, answering various questions within the same topic, following progressively complex instructions, and responding to questions based on previous responses.

\section{Data Statistics}
\begin{table}[htbp]
\resizebox{0.49\textwidth}{!}{
    \centering
    \begin{tabular}{lcc}
    \toprule
    \textbf{Topic Categories} & \textbf{Ko} & \textbf{En} \\
    \hline
    Business Studies     & 124 &  174 \\
    Chemical Biosciences  & 125 &  124 \\
    Engineering              & 125 &  139 \\
    Medical Science          & 124 & 111 \\
    Earth \& Life Sciences & 125 & 130 \\
    Physics \& Mathematics   & 118 & 149 \\
    Socio-Professional Studies    & 124 & 146 \\ 
    Liberal Arts \& Social Sciences    & 150 & 150 \\
    \hline
    Total                    & 1,015 & 1,123 \\
    \bottomrule
    \end{tabular}
    }
    \caption{Data statistics for topic categories}
    \label{tab:cate_statistics}
\end{table}

\begin{table}[htbp]
    \centering
    \begin{tabular}{lcc}
    \toprule
    \textbf{Problem Type} & \textbf{Ko} & \textbf{En} \\
    \hline
    Summarization      & 1,004 &  1,108 \\
    Multiple choice    & 1,010 &  1,048 \\
    Multiple selection & 1,003 &  1,056 \\
    Short answer       & 1,006 &  1,027 \\
    Boolean            & 1,008 &  1,070 \\
    \hline
    Total           & 5,031 &  5,309 \\
    \bottomrule
    \end{tabular}
    \caption{Data statistics for question types}
    \label{tab:data_statistics}
\end{table}

\begin{table}[htbp]
\resizebox{0.49\textwidth}{!}{
    \centering
    \begin{tabular}{lcc}
    \toprule
    \textbf{Topic Categories} & \textbf{Ko} & \textbf{En} \\
    \hline
    Business Studies     & 13 &  13 \\
    Chemical Biosciences  & 12 &  12 \\
    Engineering              & 12 &  12 \\
    Medical Science          & 12 & 13 \\
    Earth \& Life Sciences & 12 & 12 \\
    Physics \& Mathematics   & 13 & 13 \\
    Socio-Professional Studies    & 12 & 12 \\ 
    Liberal Arts \& Social Sciences    & 14 & 13 \\
    \hline
    Total                    & 100 & 100 \\
    \bottomrule
    \end{tabular}
    }
    \caption{Bilingual Data statistics for topic categories}
    \label{tab:cate_statistics-bilingual}
\end{table}

\begin{table}[htbp]
    \centering
    \begin{tabular}{lcc}
    \toprule
    \textbf{Problem Type} & \textbf{Ko} & \textbf{En} \\
    \hline
    Summarization      & 99 &  93 \\
    Multiple choice    & 99 &  87 \\
    Multiple selection & 98 &  88 \\
    Short answer       & 100 &  86 \\
    Boolean            & 99 &  92 \\
    \hline
    Total           & 495 & 446 \\
    \bottomrule
    \end{tabular}
    \caption{Bilingual Data statistics for question types}
    \label{tab:data_statistics-bilingual}
\end{table}

\noindent
Table~\ref{tab:cate_statistics} presents the number of questions for each academic category, and Table~\ref{tab:data_statistics} provides the distribution of question types. The benchmark was designed to maintain an even distribution of the five question formats, which include summarization, short answer, multiple choice, multiple selection, and true or false, across all academic domains.

We design direct bilingual evaluation dataset 9.1\% from all dataset and the statistics can be shown in Table~\ref{tab:cate_statistics-bilingual} and Table~\ref{tab:data_statistics-bilingual}. 

\begin{figure*}[!t]
    \centering
    \includegraphics[width=\linewidth]{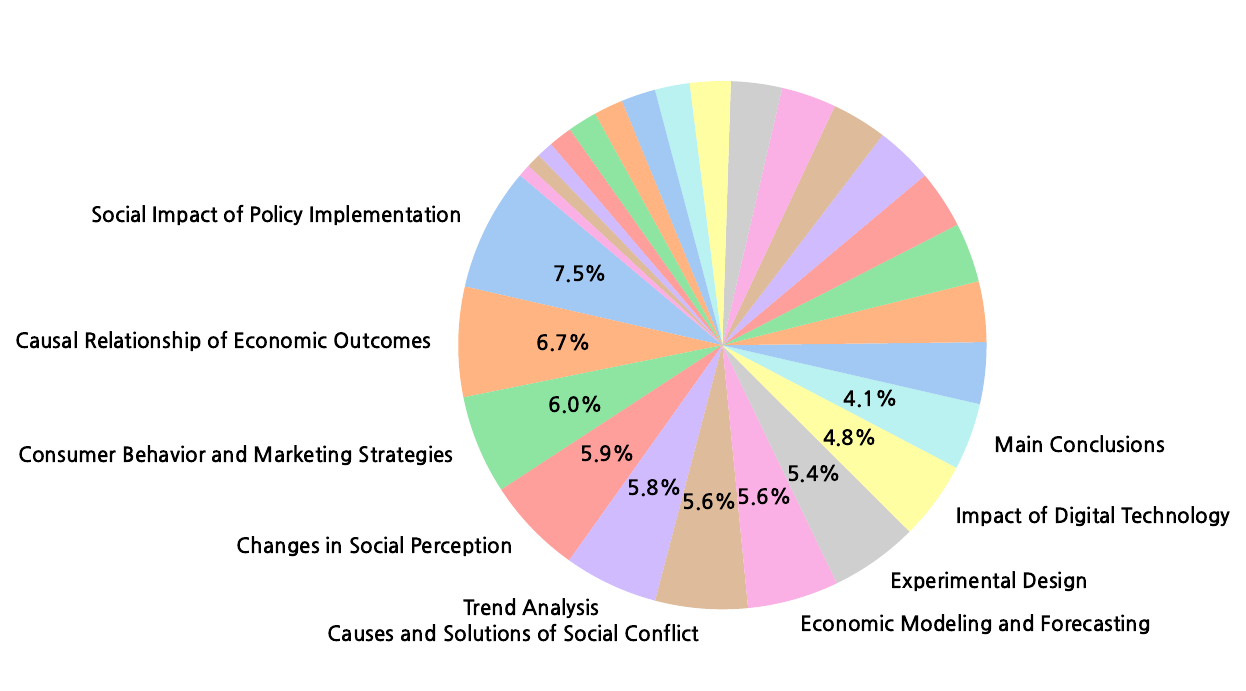}
    \caption{Frequency distribution of 27 attributes commonly shared across all question types. The balanced distribution without overconcentration on specific attributes suggests that the benchmark enables fair model evaluation across a diverse range of attributes.}
    \label{fig:pi_chart}
\end{figure*}
\subsection{Domain Diversity} 
\noindent
Figure~\ref{fig:pi_chart} illustrates the distribution of question attributes that are commonly shared across all task types, as defined in Section~\ref{sec:data_def}. A statistical analysis of their frequency reveals a mean of 177.7, a standard deviation of 89.3, and a coefficient of variation of 0.50. These results suggest that the attributes are evenly distributed, without excessive concentration in specific task types. Such a balanced distribution indicates that \texttt{ScholarBench} enables fair and comprehensive model evaluation across a diverse range of attributes.

\subsection{Query Length Distribution}
\begin{figure*}[!t]
    \centering
    \includegraphics[width=\linewidth]{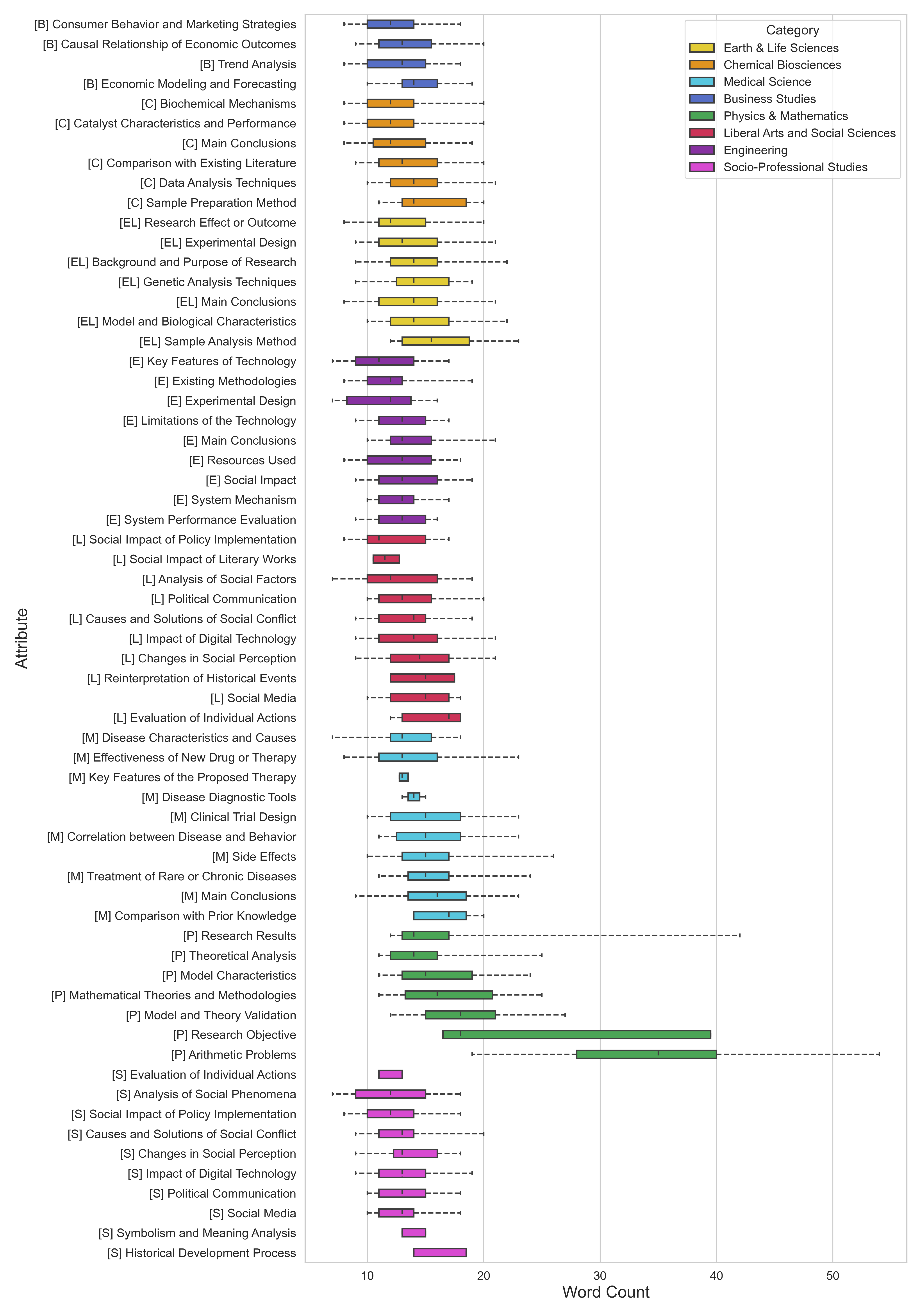}
    \caption{Distribution of query length for each attribute in English categories. The initials preceding each attribute represent abbreviations of the corresponding categories, as follows: [B] Business Studies, [C] Chemical Biosciences, [E] Engineering, [EL] Earth \& Life Sciences, [L] Liberal Arts and Social Sciences, [M] Medical Science, [P] Physics \& Mathematics, and [S] Socio-Professional Studies.}
    \label{fig:query_len_en}
\end{figure*}

\begin{figure*}[!t]
    \centering
    \includegraphics[width=\linewidth]{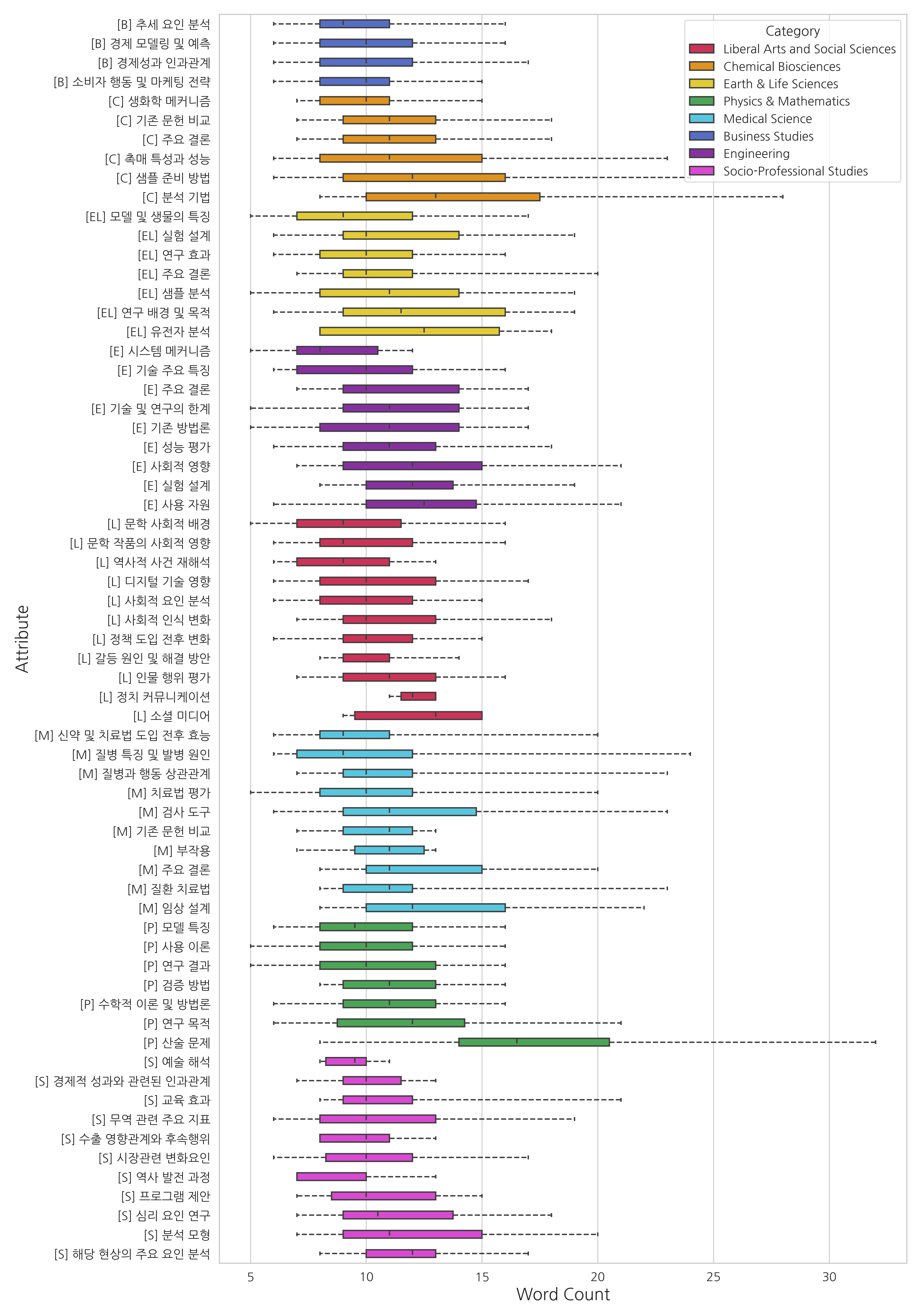}
    \caption{Distribution of query length for each attribute in Korean categories.}
    \label{fig:query_len_ko}
\end{figure*}
\noindent
We show query word length distributions for each attribute of eight domains in Figure~\ref{fig:query_len_en} and~\ref{fig:query_len_ko}.
Queries contain 10-20 words, though some instances exceed 30 words depending on the academic domain. Even when domains share the same attribute, query distributions are similar.
However, variations in query length are also observed across different attributes.
Specific attributes such as \textit{Consumer Behavior and Marketing Strategies}, \textit{Key Features of *}, \textit{Existing Methodologies}, and \textit{Disease Diagnostic Tools} tend to have shorter queries.
In contrast, attributes such as \textit{Research related}, \textit{Results of *}, and \textit{Problem of *} show a wider range of lengths.
This is considered a reflection of the unique characteristics and research topics pertinent to each academic domain, as captured by the query attributes.

The Korean dataset in Figure~\ref{fig:query_len_ko} also shows a similar pattern.

\subsection{Paragraph and Summary Length Distribution}
\begin{figure*}[!t]
    \centering
    \includegraphics[width=\linewidth]{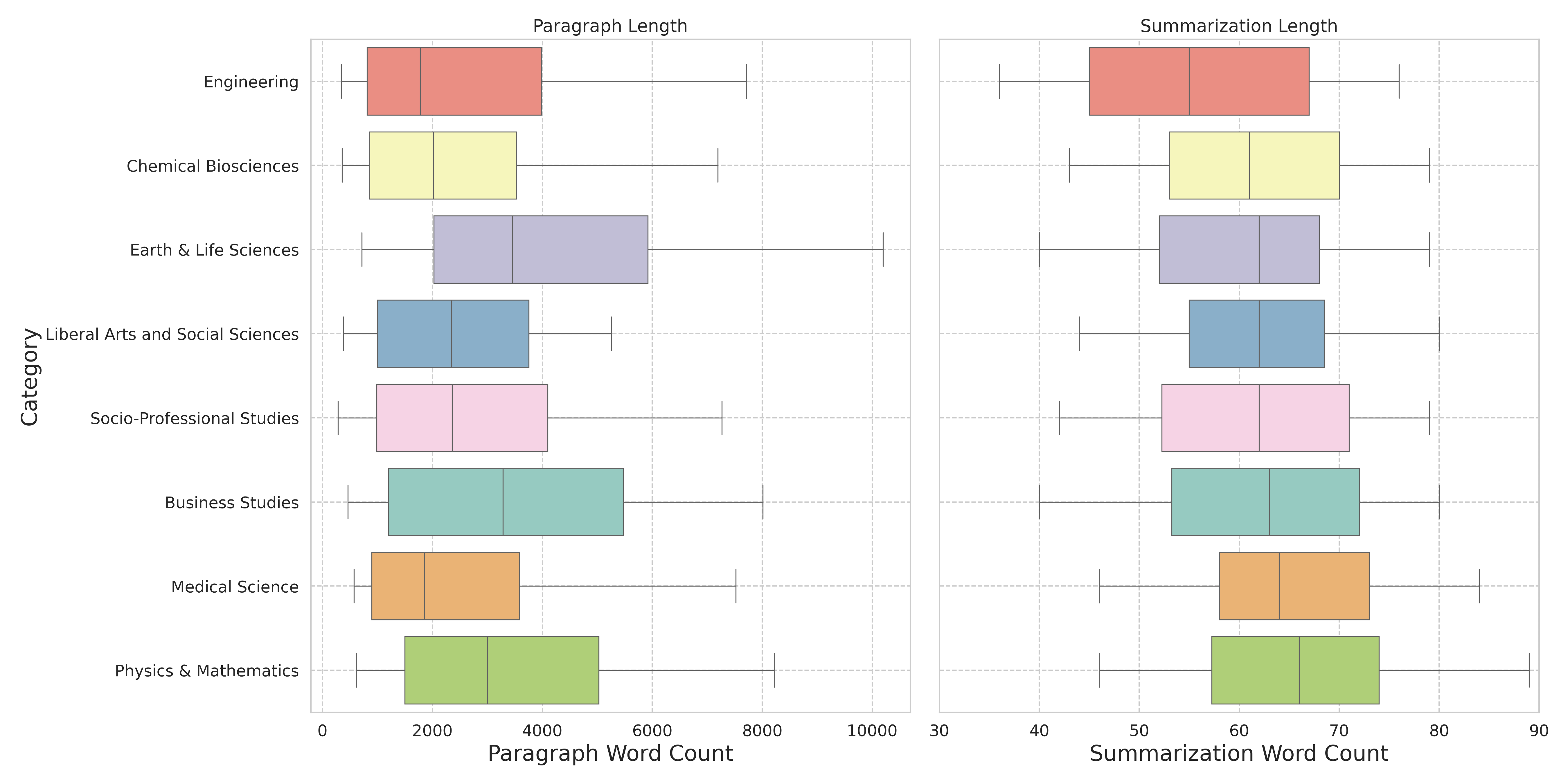}
    \caption{Length distribution of paragraph and summary for each domain.}
    \label{fig:para_sum_len}
\end{figure*}
 
\noindent
We show word length distributions for paragraphs and summaries for each domain in Figure \ref{fig:para_sum_len}.
Overall, paragraph lengths vary across academic domains and exhibit a wide distribution range.
For instance, paragraphs within the \textit{Business Studies} and \textit{Earth \& Life Sciences} domains show a relatively longer median and a wider interquartile range (IQR, representing the box length), indicating significant variance in paragraph length.
In contrast, fields such as \textit{Chemical Biosciences} and \textit{Medical Science} tend to have comparatively shorter paragraphs with a more condensed distribution.
It can also be observed that each domain includes very long paragraphs, resulting in extended whiskers in the plots.

In contrast to paragraph lengths, summary lengths demonstrate a consistent and uniform distribution across all academic domains. 
Most summaries contain between 50 and 80 words, with the median concentrating around 60-65 words in every domain. 
The interquartile range for summary lengths is also remarkably narrower compared to paragraph lengths, indicating effective length control during summary generation. 
This suggests that the summarization task encourages the generation of consistent-length summaries despite varying input lengths.

\begin{table}[!t]
    \centering
    \resizebox{0.49\textwidth}{!}{
    \begin{tabular}{ll}
        \toprule
        \bf Model & \bf Model Name \\ \hline
        llama-70b & meta-llama/Llama-3.3-70B-Instruct \\
        llama-8b & meta-llama/Llama-3.1-8B-Instruct \\
        Mistral-24b & mistralai/Mistral-Small-24B-Instruct-2501 \\
        Mistral-8b & mistralai/Ministral-8B-Instruct-2410 \\
        Qwen-72b & Qwen/Qwen2.5-72B-Instruct \\
        Qwen-32b-reasoning & Qwen/QwQ-32B \\
        Qwen-7b & Qwen/Qwen2.5-7B-Instruct \\
        Gemma2-27b & google/gemma-2-27b-it \\
        Gemma2-9b & google/gemma-2-9b-it \\
        Bllossom-70b & Bllossom/llama-3-Korean-Bllossom-70B \\
        Bllossom-8b & MLP-KTLim/llama-3-Korean-Bllossom-8B \\
        Exaone-32b-reasoning & LGAI-EXAONE/EXAONE-Deep-32B \\
        Exaone-32b & LGAI-EXAONE/EXAONE-3.5-32B-Instruct \\
        Exaone-8b & LGAI-EXAONE/EXAONE-3.5-7.8B-Instruct \\
        \bottomrule
    \end{tabular}
    }
    \caption{Model cards}
    \label{tab:ver}
\end{table}

\begin{table}[!t]
    \centering
    \resizebox{0.49\textwidth}{!}{
    \begin{tabular}{ll}
        \toprule
        \bf Model & \bf Model Name \\ \hline
        A.X-4.0 & skt/A.X-4.0-Light \\
        A.X-3.1 & skt/A.X-4.0-Light \\
        Smoothie-Qwen-7B & dnotitia/Smoothie-Qwen2.5-7B-Instruct \\
        Qwen3-8B & Qwen/Qwen3-8B \\
        Midm-2.0 & K-intelligence/Midm-2.0-Mini-Instruct \\
        kanana-8b & Qwen/Qwen2.5-72B-Instruct \\
        DNA-14B & dnotitia/DNA-2.0-14B \\
        Tri-7B & trillionlabs/Tri-7B \\
        VARCO-8B & NCSOFT/Llama-VARCO-8B-Instruct \\
        Koni-8B & KISTI-KONI/KONI-Llama3.1-8B-R-20250831 \\
        \bottomrule
    \end{tabular}
    }
    \caption{Model cards for additional evaluation (Table .\ref{tab:additional_exp_model_results}) }
    \label{tab:ver2}
\end{table}
\section{Applied Hyperparameters}
\subsection{Model Cards}
\label{appx:model-card}
We show model cards used in experiments of this paper in Table~\ref{tab:ver}.
We conducts additional experiments with Korean LLMs, and the model card used in these experiments is summarized in Table~\ref{tab:additional_exp_model_results}.

\subsection{Inference Hyperparameters}
\begin{table}[htbp]
    \centering
    \begin{tabular}{l|c}
        \toprule
        \textbf{Parameter}       & \textbf{Value} \\
        \midrule
        TensorParallelSize       & 4              \\
        DType                    & bfloat16       \\
        GpuMemoryUtilization     & 0.95           \\
        Seed                     & 42             \\
        Temperature              & 0              \\
        MaxTokens                & 32k            \\
        TopK                     & 1              \\
        DoSample                 & False          \\
        BatchSize                & 1              \\
        \bottomrule
    \end{tabular}
    \caption{vLLM Inference Hyperparameters}
    \label{tab:vllm_hparams}
\end{table}

\noindent
We conducted inference using vLLM to support various experiments and evaluations. For a fair comparison across models, all hyperparameters were uniformly set to commonly used default values. To ensure reproducibility and consistency in model outputs, we set the temperature to 0, enabling greedy decoding. Furthermore, to prevent response loss and performance degradation often observed with large batch processing in the vLLM environment, and to maintain consistent output quality, we fixed the batch size to 1. The detailed hyperparameter settings are summarized in Table~\ref{tab:vllm_hparams}.

\section{Evaluation Metrics}
\label{appx:metrics}
\begin{itemize}
    \item 
\textbf{BERTScore}~\cite{zhang2020BERTScore} computes semantic similarity between a generated sequence and a reference sequence by taking the maximum similarity between each token in one sequence and the tokens in the other. We use BERTScore to assess semantic alignment between model-generated answers and reference texts:
$$
    \text{BERTScore} = \frac{1}{N} \sum_{i=1}^{N} \max_{j} \cos ( \mathbf{h}_{i}^{\mathrm{Ref}}, \mathbf{h}_{j}^{\mathrm{Gen}} )
$$
    \item 
\textbf{BLEURT}~\cite{bleurt} evaluates the semantic similarity between the generated and reference texts using a pre-trained language model. Unlike BLEURT, BLEU~\cite{bleu} measures n-gram precision between the generated and reference texts, focusing on how much of the generated content is contained within the reference.
    \item 
\textbf{ROUGE}~\cite{lin-2004-rouge} evaluates n-gram recall, assessing how much of the reference content is covered in the generated text. Variants such as ROUGE-1, ROUGE-2, and ROUGE-L measure different aspects of textual overlap. ROUGE is particularly well-suited for evaluating summarization, as it reflects how well the generated summary captures key information from the source text.
\end{itemize}

\section{Additional Analysis}
\subsection{Query Embedding Analysis}
\begin{figure*}[!t]
    \centering
    \includegraphics[width=\linewidth]{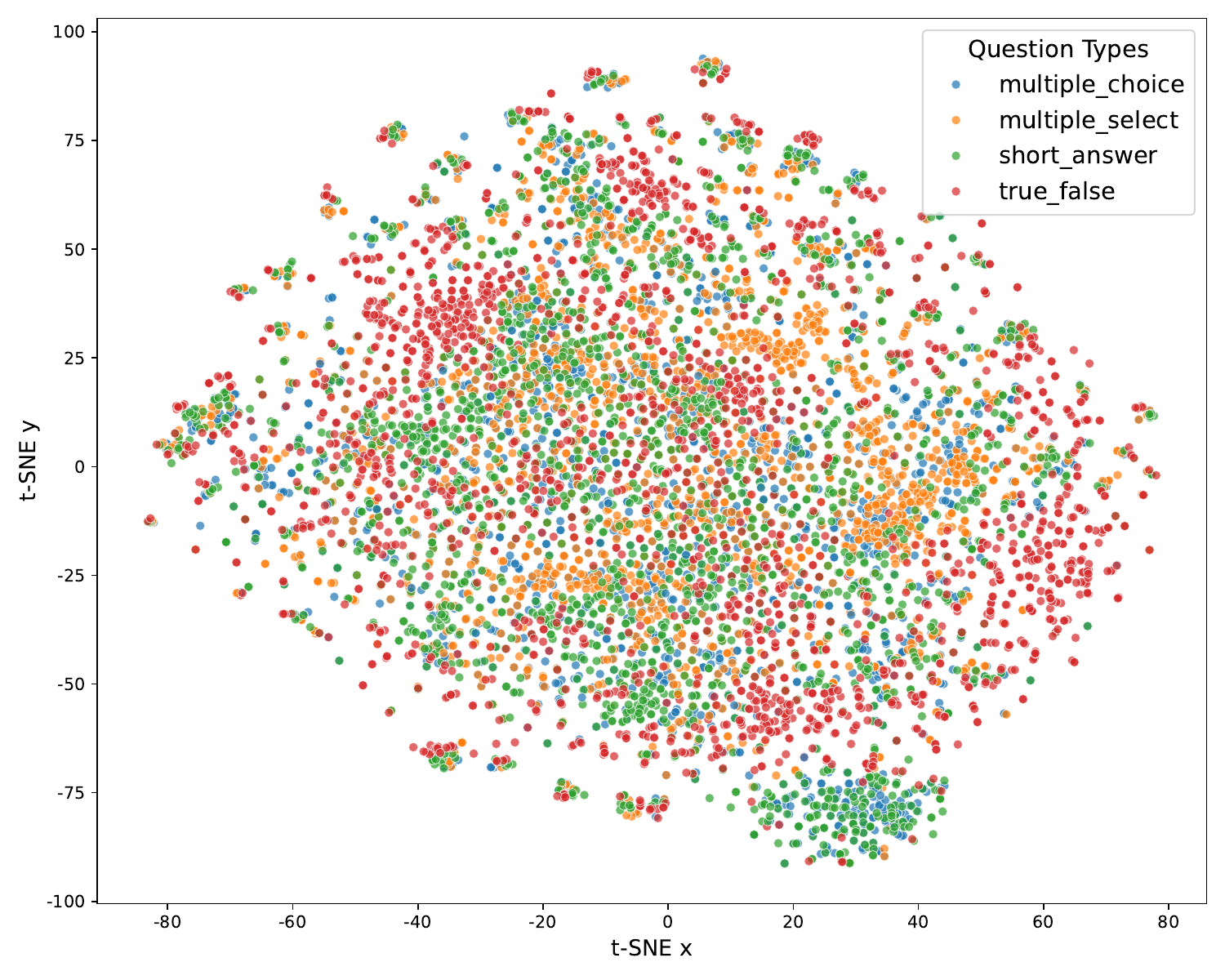}
    \caption{t-SNE visualization of query embeddings in Scholar Bench. The wide and overlapping distribution across question types suggests that the queries are semantically diverse and not bound to specific task types, enabling fairer evaluation.}
    \label{fig:t_sne}
\end{figure*}
\noindent
To validate the semantic diversity of the proposed \texttt{ScholarBench}, we conducted a t-SNE-based visualization analysis of the query embeddings~\cite{cai2022theoreticalfoundationstsnevisualizing}. Since \texttt{ScholarBench} is a bilingual dataset comprising both Korean and English, it is essential that semantically similar sentences are mapped to a shared embedding space regardless of language. To this end, we employed LaBSE~\cite{feng-etal-2022-language} as the sentence encoder, which is trained to map semantically equivalent sentences across multiple languages into the same vector space.

Figure~\ref{fig:t_sne} presents the result of applying t-SNE to the query embeddings from \texttt{ScholarBench}. Notably, the embeddings are widely dispersed without forming clusters based on question type. This suggests that the queries exhibit rich semantic variation and are not restricted to specific task formats, thereby enabling fairer and more comprehensive evaluation of language models.

\subsection{Category Analysis}
\label{appx:cate-anal}
\noindent
To analyze the characteristics that LLMs have within each domain, Figure~\ref{fig:cate_eval} visualizes the performance of the top-10 models from the main experiments for the eight academic domains.
First, in the summarization (R-1) task, most models demonstrate high scores in the Engineering and Medical Science domain, whereas they exhibit relatively lower scores in Economy \& Management and Socio-Professional Studies.
Summarization is relatively easier in structured technical texts, whereas economics and social science texts are more challenging due to contextual dependency and linguistic variability.
Despite mild overall variation, model performance is consistently lower in the Economy \& Management and Liberal Arts domains in the short-answer task.
The Multiple Selection (A-4) task shows large variance in model performance. Models achieve high accuracy in Biology \& Earth Science, but perform poorly on categories that require abstract or complex reasoning, such as Liberal Arts and Physics \& Mathematics.
Model reasoning ability in multiple-selection questions varies by category. For both MCQ and boolean formats, performance is higher in Engineering and Chemical \& Biochemistry, but lower in Biology, Liberal Arts, and Physics. 
The Qwen-32b-reasoning model performs particularly well in reasoning-focused domains, likely due to its pretraining objectives. In contrast, o3-mini and Mistral-24b show stable performance across domains, reflecting stronger generalization.

According to these results, the average performance of models is insufficient to explain domain-specific performance characteristics. 
Thus, both domain generalization ability and domain-specific reasoning ability should be considered in LLM evaluation. 
Fine-grained benchmarks like \texttt{ScholarBench} are effective in quantitatively revealing these imbalances in performance distribution.

\subsection{Short Answer Bilingual Results}
\begin{table}[t]
\centering
\resizebox{0.49\textwidth}{!}{
\begin{tabular}{lcc|lcc}
\toprule
\multicolumn{3}{c|}{\textbf{English}} & \multicolumn{3}{c}{\textbf{Korean}} \\
\textbf{Model} & \textbf{BERTScore} & \textbf{BLEURT} & \textbf{Model} & \textbf{BERTScore} & \textbf{BLEURT} \\
\midrule
o3-mini              & 0.852 & 0.328 & o3-mini              & 0.868 & 0.403 \\
GPT-4o               & 0.851 & 0.342 & GPT-4o               & 0.864 & 0.367 \\
o1-mini              & 0.851 & 0.334 & o1-mini              & 0.863 & 0.362 \\
llama-70b            & 0.850 & 0.311 & llama-70b            & 0.866 & 0.359 \\
Qwen-72b             & 0.852 & 0.337 & Qwen-72b             & 0.867 & 0.375 \\
Qwen-32b-r   & 0.846 & 0.305 & Qwen-32b-r   & 0.861 & 0.354 \\
Mistral-24b          & 0.847 & 0.313 & Mistral-24b          & 0.859 & 0.335 \\
Gemma2-27b           & 0.847 & 0.299 & Gemma2-27b           & 0.864 & 0.347 \\
Gemma2-9b            & 0.847 & 0.307 & Gemma2-9b            & 0.855 & 0.318 \\
Qwen-7b              & 0.846 & 0.286 & Qwen-7b              & 0.848 & 0.279 \\
\bottomrule
\end{tabular}
}
\caption{Short answer evaluation results on English and Korean. Qwen-32b-r is Qwen-32b-reasoning.}
\label{tab:sa_anal}
\end{table}

\noindent
Table~\ref{tab:sa_anal} presents the short answer performance of 10 selected models for both Korean and English, with model selection informed by the results detailed in Table~\ref{tab:main-result}. 
Overall, the BERTScore across the two languages is largely similar, exhibiting only minor differences that are likely attributable to inherent linguistic characteristics. 
Beyond BERTScore, which quantifies semantic similarity, BLEURT was also utilized for evaluation. Notably, the linguistic discrepancy between the two languages appears more pronounced when assessed using BLEURT.

\subsection{Analysis for Reasoning Ability}

\begin{table*}[t]
\centering
\small
\resizebox{\textwidth}{!}{
\begin{tabular}{p{0.32\linewidth} p{0.2\linewidth} p{0.40\linewidth}}
\toprule
\textbf{Passage (excerpt)} \\
\midrule
\multicolumn{3}{p{\linewidth}}{
\textit{Dominican youth of Haitian descent face significant barriers to education due to lack of documentation, societal prejudice, and economic hardship. Many children receive Dominican birth certificates, only to have them cancelled arbitrarily. High dropout rates are observed, especially for females, who may encounter additional challenges from traditional gender roles. Despite these adversities, some students continue their education.}
} \\
\midrule
\textbf{Question} & \textbf{Answer} & \textbf{Clue} \\
\midrule
What specific challenge related to documentation impacts school participation for Dominican females of Haitian descent? & Lack of documentation & [C1] Many children receive Dominican birth certificates, [C2] only to have them cancelled arbitrarily \\ \midrule
What factors contribute to the educational challenges faced by Dominican girls of Haitian descent? (Select all that apply) & a) Cultural attitudes like machismo, b) Economic hardship & [C1] High dropout rates are observed, especially for females, [C2] who may encounter additional challenges from traditional gender roles. [C3] Dominican youth of Haitian descent face significant barriers to education due to lack of documentation, societal prejudice, and economic hardship  \\ \midrule
What impact can the lack of documentation have on youth education? & a) Denial of access to national exams & [C1] Dominican youth of Haitian descent face significant barriers to education due to lack of documentation. \\ \midrule
The absence of documentation does not affect the educational success of Dominican females of Haitian descent. (True/False) & False & [C1] Dominican youth of Haitian descent face significant barriers to education due to lack of documentation. [C2] High dropout rates are observed, especially for females. \\
\bottomrule
\end{tabular}
}
\caption{Step-by-step analysis of LLMs' contextual reasoning. [C] represents a clue for generating an answer, and it is generated in the order of LLMs' reasoning. See Table~\ref{tab:reasoning-example}.}
\label{tab:reasoning-w-clue}
\end{table*}
\noindent

In this paper, we construct passages that are paired with problems to evaluate the reasoning ability of LLM. This assumes that when solving a problem using passages, LLM will reason the information in the passages to generate the correct answer. In order to experimentally verify our hypothesis, we sequentially generate the clue information used for reasoning when generating the correct answer through API-based LLM, as shown in Table~\ref{tab:reasoning-w-clue}, and show the results.

These clues are not simple information retrieval, but are used as the basis for the inference step for deriving the correct answer. Each correct answer is based on an expression directly quoted in the passage, but it is based on contextual interpretation and causal inference rather than word-level agreement. Through the generated clue information, we can see that LLM performs various cause identification. In cases where multiple correct answers are required, such as the second question, LLM shows that it integrates multiple elements to make inferences. This can be interpreted as understanding the interaction between multidimensional social factors, not simple pattern recognition.

\subsection{Analysis of Summary Length and Performance}

\begin{figure*}[!t]
    \centering
    \includegraphics[width=\linewidth]{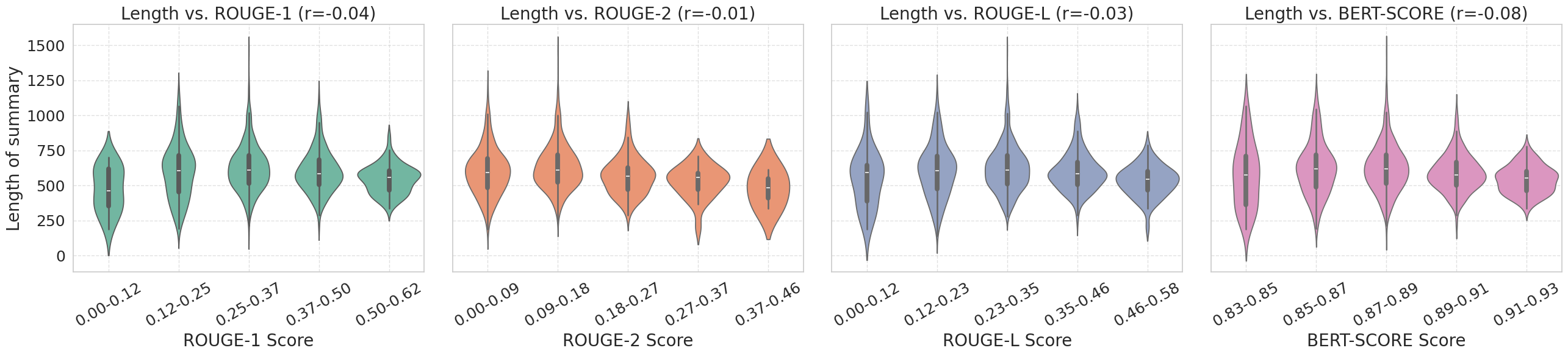}
    \vspace{0.5em} 
    \includegraphics[width=\linewidth]{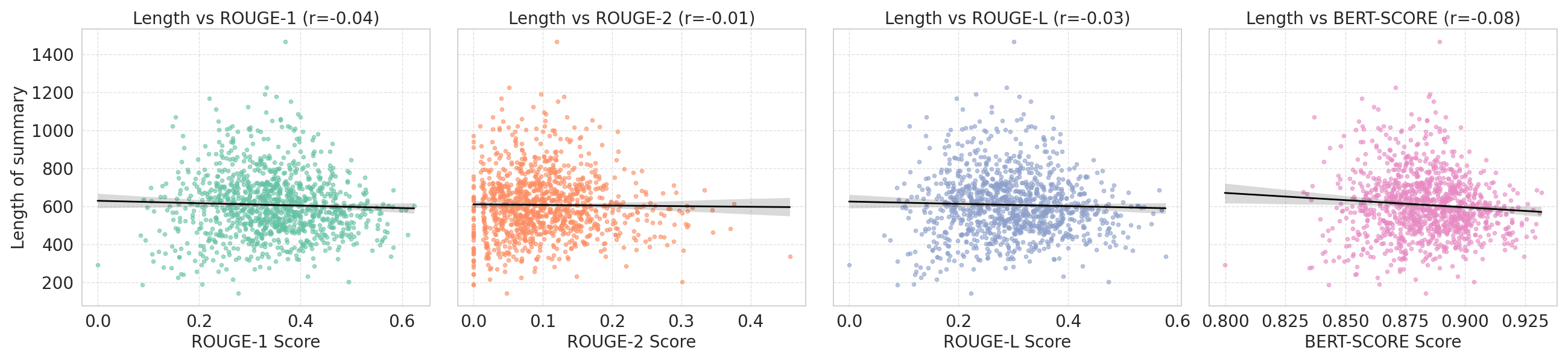}
    \caption{Length distribution of summaries and regression analysis for English.}
    \label{fig:len_dist_sum_en}
\end{figure*}

\begin{figure*}[!t]
    \centering
    \includegraphics[width=\linewidth]{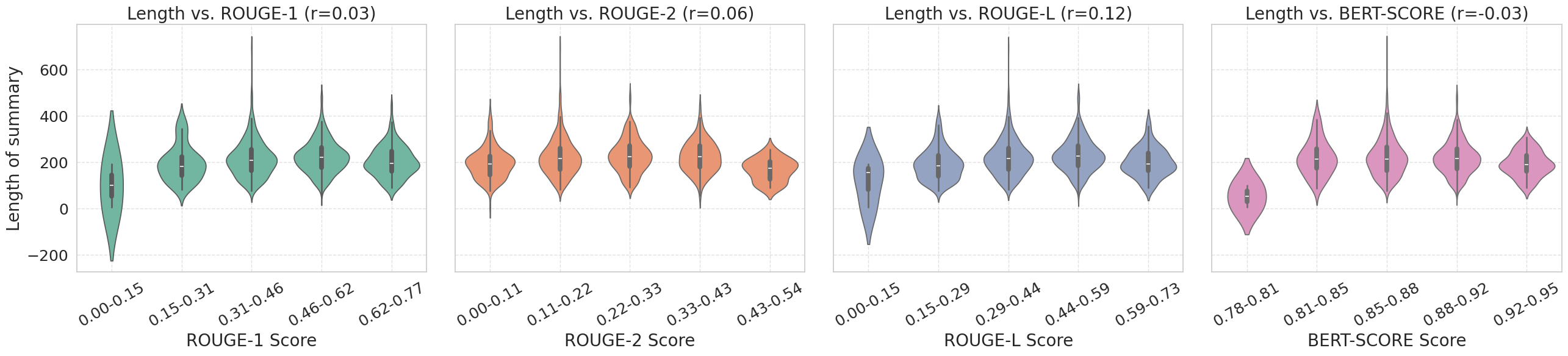}
    \vspace{0.5em} 
    \includegraphics[width=\linewidth]{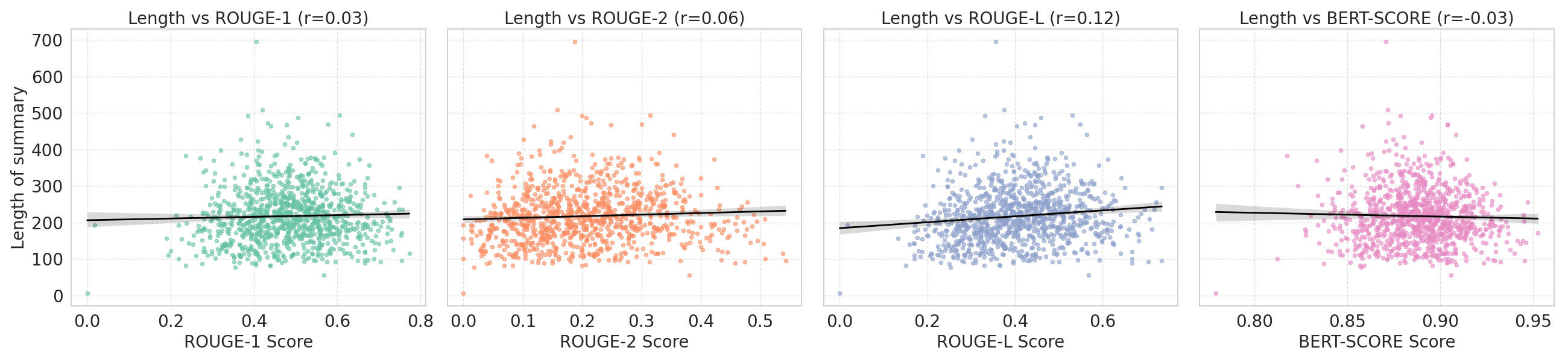}
    \caption{Length distribution of summaries and regression analysis for Korean.}
    \label{fig:len_dist_sum_ko}
\end{figure*}
\noindent
Figures~\ref{fig:len_dist_sum_en} and~\ref{fig:len_dist_sum_ko} visualize the relationship between summary length and evaluation scores for English and Korean summaries, respectively. 
In each figure, the upper plots show the distribution of summary lengths across various performance score ranges, while the lower plots illustrate the regression relationship between summary scores and their corresponding lengths.
Across both languages, a low correlation is observed between evaluation metrics and summary length, suggesting that summary length does not significantly influence performance.
Furthermore, this indicates that the constructed dataset is not optimized for a particular length. 
Instead, it maintains consistent quality across a wide range of lengths according to the evaluation metrics, thus reflecting its balanced nature.

\subsection{Human Evaluation}
\begin{figure*}[!t]
    \centering
    \includegraphics[width=\linewidth]{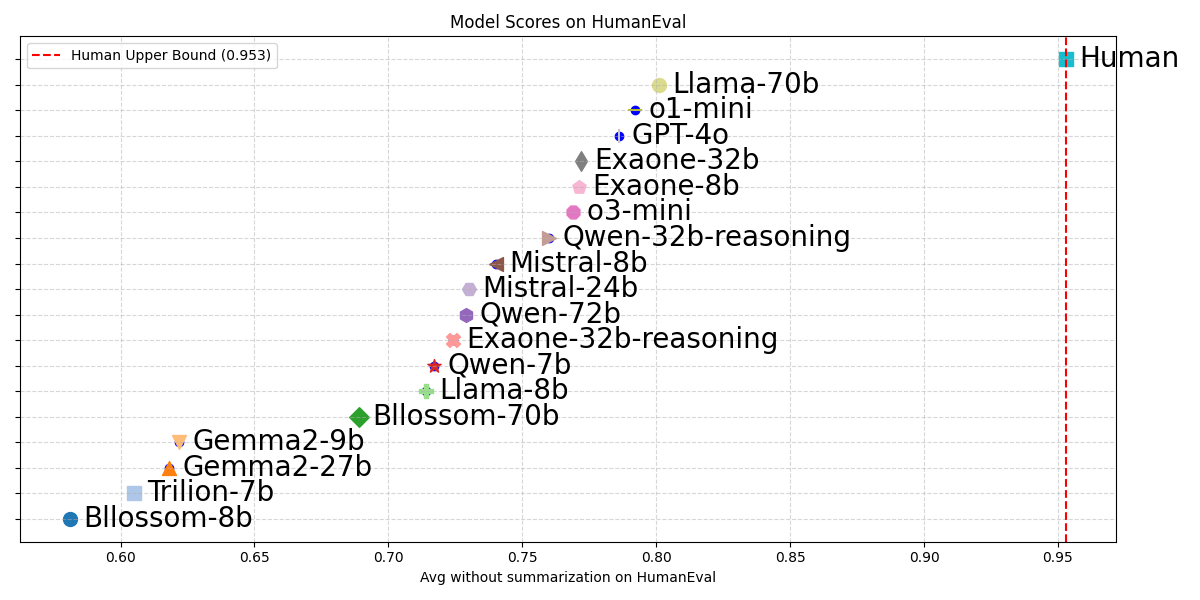}
    \caption{Comparison between human evaluation and model performance on 1\% of the data, using the same examples as in Table~\ref{tab:data-human-eval-1}. All evaluations are conducted with access to the corresponding paragraph context.}
    \label{fig:comp_human_n_models}
\end{figure*}
\begin{table}[]
\centering
\small
\resizebox{0.48\textwidth}{!}{
    \begin{tabular}{l|cc|cc|c}
    \toprule
    \multirow{2}{*}{\begin{tabular}[c|]{@{}l@{}}Language\end{tabular}} & \multicolumn{2}{c|}{Evaluation} & \multicolumn{2}{c|}{Assessment}  & \multirow{2}{*}{\begin{tabular}[c]{@{}l@{}}Kappa Coefficient\end{tabular}} \\ 
             & A & B & A & B & \\
    \midrule
    English  & 4.50 & 4.13 & 4.11 & 4.08 & 0.614 \\
    Korean  & 4.19 & 3.81 & 3.97 & 3.91 &  0.706 \\
    \bottomrule
    \end{tabular}
}
\caption{Human evaluation and data difficulty assessment using a 1–5 scale and Kappa coefficient on 1\% random data.}
\label{tab:data-human-eval-1}
\end{table}
\noindent
Figure~\ref{fig:comp_human_n_models} compares human evaluation results with the performance of various LLMs on a randomly sampled 1\% subset of the entire dataset. 
The examples used in this comparison are identical to those in Table~\ref{tab:data-human-eval-1}, and all evaluations are conducted with access to the corresponding paragraph.
Similar to the high inter-annotator agreement observed in the human evaluation (Cohen’s Kappa: 0.614 for English and 0.706 for Korean), the distribution of model performance on the sampled data shows clear differentiation.
The consistency between model predictions and human judgments further supports the \textit{reliability} of the \texttt{ScholarBench} as a quantitative evaluation standard.


In the graph, the average score from human evaluation (i.e., the Human Upper Bound) is presented as a reference line, above which most models do not reach. 
This demonstrates that \texttt{ScholarBench} is \textit{sensitive} enough to capture fine-grained differences in model performance.

\subsection{Additional Model Results}
\begin{table*}[!t]
    \centering
    \resizebox{\textwidth}{!}{
    \begin{tabular}{lcccccccccc}
    \toprule
    \multirow{2}{*}{\begin{tabular}[c]{@{}l@{}}Model\end{tabular}} & \multicolumn{3}{c}{Summarization}		& {Short Answer}            & \multicolumn{3}{c}{Multiple Selection}	& \multirow{2}{*}{\begin{tabular}[c]{@{}l@{}}MCQ\end{tabular}} & \multirow{2}{*}{\begin{tabular}[c]{@{}l@{}}Boolean\end{tabular}} & \multirow{2}{*}{\begin{tabular}[c]{@{}l@{}}Avg\end{tabular}} \\ \cmidrule(lr){2-4} \cmidrule(lr){5-5} \cmidrule(lr){6-8} 
      & R-1	& R-2	& R-L	 & BERTScore	& A-2	& A-3	& A-4	&    &         &     \\ \hline
        A.X-4.0  & 0.387 & 0.143 & 0.335 & 0.857 & 0.568 & 0.454 & 0.419 & 0.698 & 0.770 & 0.515 \\
        A.X-3.1 & 0.358 & 0.118 & 0.304 & 0.857 & 0.530 & 0.308 & 0.281 & 0.641 & 0.757 & 0.461 \\
        Smoothie-Qwen-7B  & 0.391&	0.144	&0.340	&0.847&	0.560&	0.453	&0.424&	0.698	&0.761	&0.513 \\
        Qwen3-8B & 0.385	& 0.134& 	0.330& 	0.851& 	0.533& 	0.431	& 0.390& 	0.695& 	0.746	& 0.499 \\
        Midm-2.0 & 0.403	&0.152&	0.350&	0.852&	0.555	&0.373	&0.339&	0.680	&0.615	&0.480 \\
        kanana-8b & 0.363	&0.118&	0.307&	0.854&	0.510&	0.395&	0.363&	0.640	&0.749	&0.478 \\
        DNA-14B & 0.352&	0.114&	0.303	&0.849	&0.547&	0.441&	0.412	&0.692	&0.761	&0.497 \\
        Tri-7B &0.375&	0.131&	0.324&	0.852	&0.509&	0.368&	0.342&	0.699&	0.627	&0.470 \\
        VARCO-8B & 0.278&	0.072	&0.234&	0.839&	0.564	&0.445&	0.414&	0.661&	0.546&	0.450 \\
        Koni-8B & 0.358&	0.117&	0.306&	0.836&	0.400&	0.341	&0.323&	0.599&	0.439&	0.413 \\
    \bottomrule 
    \end{tabular}
    }
    \caption{Additional overall evaluation results of \texttt{ScholarBench}, based on the 2025-09 snapshot.}
    \label{tab:additional_exp_model_results}
\end{table*}
\noindent
Table~\ref{tab:additional_exp_model_results} shows additional evaluation model results for \texttt{ScholarBench}.

\section{Prompt Template}
\label{appx:prompt-template}
\subsection{Synthetic Data Generation}
\noindent
The method for generating paragraphs is as follows. We extract paragraphs from full research papers using a sliding window approach. Specifically, we set the window size to 33\% of the full document and slide it with a 20\% overlap to generate each paragraph.

The example below illustrates the prompt used for generating synthetic data. This prompt includes attribute definitions and standards for each item, defined as criteria, alongside few-shot examples sampled from the 1st synthetic data. The prompt for generating the 1st synthetic data is applied without including few-shot examples.
\begin{instructionbox}{English data generation prompt}
    To create an evaluation set, we need to generate five types of questions and answers as follows: \\
    \{summary, short\_answer, multiple\_choice, multiple\_select, true\_false\}. \\
    $\langle$Order$\rangle$ \\
    1. Generate questions for all types except 'summary' based on the given $\langle$Topic$\rangle$. If the $\langle$Topic$\rangle$ does not match the document  content, feel free to create an appropriate topic. \\
    2. For all question types except 'summary', create questions using multi-hop reasoning. Please provide the reasoning process explaining why it is multi-hop. \\
    3. Verify that the generated questions match the provided document. \\
    4. Evaluate whether the generated answers correctly respond to the questions. \\
    5. Confirm that the generated answers are consistent with the provided document. \\
    6. Identify the question that is most similar to the $\langle$Topic$\rangle$ \\
    $\langle$Topic$\rangle$: \\
    $\langle$Format$\rangle$: \\
    $[$Summary$]$ \\  
    Write a summary. \\
    $[$short\_answer question$]$ \\  
    (Q) Write a question.  \\
    (A) Write an answer. \\
    $[$multiple\_choice question$]$ \\  
    (Q) Write a question. \\ 
    A) Write choice.  \\
    B) Write choice.  \\
    C) Write choice.  \\
    D) Write choice.  \\
    (A) Write the correct answer. \\
    $[$multiple\_select question$]$ \\  
    (Q) Write a question. \\ 
    A) Write choice.  \\
    B) Write choice.  \\
    C) Write choice.  \\
    D) Write choice.  \\
    (A) Write the correct answers. \\
    $[$true\_false question$]$  \\
    (Q) Write a question. + (True/False) \\  
    (A) Write the correct answer. \\
    \label{prompt:english-question-gen}
\end{instructionbox}

\begin{instructionbox}{Korean data generation prompt}
    평가셋을 만들기 위해,우리는 다음과 같은 다섯 가지 유형의 질문과 답변을 만들어야 합니다: \\
    \{요약, 단답형 질문, 객관식 질문, 다중선택 질문, 참/거짓 질문\}. \\
    $\langle$순서$\rangle$ \\
    1. '요약'을 제외한 유형에서 $\langle$주제$\rangle$에 맞게 질문을 생성하세요. 만약 $\langle$주제$\rangle$와 문서 내용이 일치하지 않는다면, 질문 주제를 임의로 생성하세요. \\
    2. 요약을 제외한 문제 유형 multi-hop으로 생성하세요.왜 multi-hop인지 사고 과정을 제공해주세요. \\
    3. 생성한 질문이 주어진 문서와 일치하는지 평가하세요. \\
    4. 생성한 답변이 질문에 대하여 올바른지 평가하세요. \\
    5. 생성한 답변이 주어진 문서와 일치하는지 평가하세요. \\
    6. $\langle$주제$\rangle$와 가장 유사한 질문을 알려주세요. \\
    $\langle$주제$\rangle$:  \\
    $\langle$형식$\rangle$: \\
    $[$요약$]$ \\
    요약을 작성하시오. \\
    $[$단답형 질문$]$ \\
    (Q) 질문을 작성하시오. \\
    (A) 답변을 작성하시오. \\
    $[$객관식 질문$]$ \\
    (Q) 질문을 작성하시오. \\
    A) 선택지를 작성하시오. \\
    B) 선택지를 작성하시오. \\
    C) 선택지를 작성하시오. \\
    D) 선택지를 작성하시오. \\
    (A) 정답을 작성하시오. \\
    $[$다중선택 질문$]$ \\ 
    (Q) 질문을 작성하시오. \\
    A) 선택지를 작성하시오. \\
    B) 선택지를 작성하시오. \\
    C) 선택지를 작성하시오. \\
    D) 선택지를 작성하시오. \\ 
    (A) 정답을 작성하시오. \\
    $[$참/거짓 질문$]$ \\
    (Q) 질문을 작성하시오. + (참/거짓) \\
    (A) 정답을 작성하시오. \\
    \label{prompt:korean-question-gen}
\end{instructionbox}

\subsection{Evaluation Prompt}
\begin{instructionbox}{Multiple-choice question prompt}
A multiple-choice question with a single correct answer is provided.
The Question contains the given question text.
The Choices include four answer options for the question, and you must select the most appropriate one.
fewshot1 and fewshot2 provide examples of selecting the most appropriate answer from the Choices for the given Question. \\
The question is presented in the following format: \\
Question: \{question\} \\
Choices: \{choices\} \\
fewshot1: \{fewshot1\} \\
fewshot2: \{fewshot2\} \\
Refer to the few shot examples, read the Question, and output only the letter corresponding to the correct answer from the Choices.
Do not provide any additional explanations, reasons, or detailed content.
Only output the letter of the correct answer.
\label{prompt:mc_en}
\end{instructionbox}

\begin{instructionbox}{Multiple selection question prompt}
A multiple selection question with one or more correct answers is provided.
The Question contains the given question text.
The Choices include four answer options for the question, and you must select the most appropriate one or more answers.
fewshot1 and fewshot2 provide examples of selecting the most appropriate one or more answers from the Choices for the given Question. \\
The question is presented in the following format: \\
Question: \{question\} \\
Choices: \{choices\} \\
fewshot1: \{fewshot1\} \\
fewshot2: \{fewshot2\} \\
Refer to the fewshot examples, read the Question, and output the letter(s) corresponding to the correct answer(s) from the Choices in Python list format.
Do not provide any additional explanations, reasons, or detailed content.
Only output the list of correct answer letters.
\label{prompt:ms_en}
\end{instructionbox}

\begin{instructionbox}{Short answer prompt}
A short answer question is provided.
The Question contains the given question text.
fewshot1 and fewshot2 provide examples of short-answer responses to the Question.
The question is presented in the following format: \\
Question: \{question\} \\
fewshot1: \{fewshot1\} \\
fewshot2: \{fewshot2\} \\
Refer to the fewshot examples, read the Question, and provide a short-answer response.
Answer only with keywords or short phrases.
Do not use complete sentences or provide additional details or explanations.
Only output the correct answer.
\label{prompt:sa_en}
\end{instructionbox}

\begin{instructionbox}{Boolean}
A True or False question is provided, where the correct answer is either 0 or 1.
The Question contains the given question text.
fewshot1 and fewshot2 provide examples of determining whether the Question is true or false.
The question is presented in the following format: \\
Question: \{question\} \\
fewshot1: \{fewshot1\} \\
fewshot2: \{fewshot2\} \\
Refer to the fewshot examples, read the Question, and determine whether it is true or false.
Output 1 if true and 0 if false.
Do not provide any additional explanations, reasons, or details.
Only output the corresponding number.
\label{prompt:bool_en}
\end{instructionbox}

\begin{instructionbox}{Summarization}
A paragraph is provided.
The Paragraph is the text to be summarized.
fewshot1 and fewshot2 provide examples of creating a simple and clear summary of the given paragraph.
Read the following Paragraph and provide a brief and clear summary. 
Output only the summary. \\
Paragraph: \{paragraph\} \\
fewshot1: \{fewshot1\} \\
fewshot2: \{fewshot2\}
\label{prompt:summary_en}
\end{instructionbox}
\noindent
The evaluation prompts used in this paper are as follows: multiple choice question~\ref{prompt:mc_en}, multiple selection question~\ref{prompt:ms_en}, short answer~\ref{prompt:sa_en}, boolean~\ref{prompt:bool_en}, and summarization~\ref{prompt:summary_en}.

\subsection{Example of Benchmark Construction}

\begin{table*}[htbp]
    \renewcommand{\arraystretch}{0.9} 
    \begin{center}
    \begin{tabular}{|p{0.95\linewidth}|} 
    \hline
    \rowcolor{gray!10}\textbf{Step-1: 1st synthetic data} \\
    \hline
    question: Which industries have experienced significant advancements due to GPT applications?  \\
    answer: Education and healthcare? \\
    \hline
    question: What are some of the main technologies used in GPT models? (Select all that apply) \\
      choices: [
        a) Deep learning models,
        b) Rule-based algorithms,
        c) Transformer architecture,
        d) Manual data labeling
    ] \\
  answer: [
    a,
    c
    ] \\
    \hline
    question: What technology does GPT mainly utilize?\\
    choices: [
        a) Rule-based systems,
        b) Deep learning models,
        c) Supervised learning,
        d) Semantic analysis
        ] \\
    answer: b \\
    \hline
    question: GPT only follows hand-coded rules to generate text. \\
    answer: False  \\
    \hline
    \rowcolor{gray!10}\textbf{Step-2: 2nd synthetic data} \\
    \hline
    question: Which industry has been notably transformed by GPT through personalized learning and automation of academic tasks?  \\
    answer: Education \\
    \hline
    question: Based on the principles underlying GPT, which technologies contribute to its ability to generalize language patterns beyond memorization? (Select all that apply) \\
      choices: [
        a) Pretraining on large-scale textual data,
        b) Rule-based decision trees,
        c) Attention-based neural networks,
        d) Explicit grammar rules
      ] \\
      answer: [
        a,
        c
      ] \\
    \hline
    question: Based on GPT’s ability to generate human-like text through large-scale training data and neural architectures, which underlying technology enables this capability?  \\
    choices: [
      a) Rule-based systems,
      b) Deep learning models,
      c) Supervised learning,
      d) Semantic analysis
    ] \\
    answer: b \\
    \hline
    question: GPT operates based on predefined rules rather than learning from data patterns. \\
    answer: False \\
    \hline
    \rowcolor{gray!10}\textbf{Step-3: Human annotation} \\
    \hline
    question: Which industry, often associated with knowledge transfer and learning, has seen significant transformation through GPT-driven innovations such as intelligent tutoring systems and automated content generation?  \\
    answer: Education \\
    \hline
    question: Considering GPT’s architecture and learning process, which of the following elements enable it to generate semantically coherent and contextually relevant responses by leveraging hierarchical representations of language? \\
      choices: [
        a) Transformer-based deep neural networks,
        b) Unsupervised pretraining on diverse corpora,
        c) Rule encoding for language syntax,
        d) Self-attention mechanisms enabling contextual word representation
      ] \\
      answer: [
        a,
        b,
        d
      ] \\
    \hline
    question: Considering that GPT generates coherent text by learning statistical patterns from large datasets using multi-layered neural networks, which of the following best describes the core technology it is built upon? \\
    choices: [
      a) Rule-based systems,
      b) Deep learning models,
      c) Supervised learning,
      d) Semantic analysis
    ] \\
    answer: b \\
    \hline
    question: GPT produces human-like language by identifying statistical patterns in large-scale data rather than depending on predefined rule sets. \\
    answer: False \\
    \hline
    \end{tabular}
    \end{center}
    \caption{Example of step-by-step data generation process for Enlgish.}
    \label{tab:data_proc_ex_en}
\end{table*}

\begin{table*}[htbp]
    \begin{center}
    \begin{tabular}{|p{0.95\linewidth}|} 
    \hline
    \rowcolor{gray!10}\textbf{Step-1: 1st synthetic data} \\
    \hline
    question: 군나르 뮈르달이 구축한 발전경제학 모델의 두 가지 주요 특징은 무엇인가? \\
    answer: 사회민주주의적, 사회공학적 \\
    \hline
    question: 뮈르달 발전경제학의 한계로 지적된 요소를 모두 고르시오. \\
    choices:  [a) 지나치게 구체적인 정책 중심의 접근, b) 추상적이고 일반적인 수준에 머문 이론, c) 사회공학적 전제의 지속, d) 발전 자체에 대한 근본적 반성의 결여 ] \\
    answer: [ b, c, d ] \\
    \hline
    question: 뮈르달의 발전경제학 접근법이 갖는 주요 한계점은 무엇인가?\\
    choices: [
        a) 미국적 자본집약형 모델을 과도하게 강조했다,
        b) 구체적 맥락의 복잡성에 적용하기에 너무 일반적이고 추상적이었다,
        c) 발전을 단순한 거시적 지표의 성장으로만 보았다,
        d) 문화상대주의적 접근을 완전히 배제했다
        ] \\
    answer: b \\
    \hline
    question: 뮈르달의 발전경제학은 사회공학적 전제를 완전히 제거했다. (참/거짓)\\
    answer: 거짓\\
    \hline
    \rowcolor{gray!10}\textbf{Step-2: 2nd synthetic data} \\
    \hline
    question: 군나르 뮈르달의 주요 경제 모델은 무엇인가? \\
    answer: 사회민주주의 \\
    \hline
    question: 군나르 뮈르달의 발전경제학이 접근하는 방식의 주요 특징은 무엇인가요? (모두 선택) \\
    choices: [ a) 거시적 지표의 성장, b) 전 사회적인 변화와의 결합, c) 사회공학적 접근, d) 정책 도입 전후의 사회적 영향 분석 ] \\
    answer: [ b, c ] \\
    \hline
    question: 군나르 뮈르달의 발전경제학 접근법에서 강력한 중앙의 권위에 근거한 발전을 상정할 수 밖에 없는 이유는 무엇인가? \\
    choices: [
            a) 사회적 변화의 복잡성,
            b) 사회공학적 접근의 고수,
            c) 근대화 이론에 대한 반대,
            d) 미국 주도의 자본주의 질서
        ] \\
    answer: a \\
    \hline
    question: 군나르 뮈르달의 발전경제학은 단순한 경제 지표의 성장에 초점을 맞추고 있다. (참/거짓)\\
    answer: 거짓 \\
    \hline
    \rowcolor{gray!10}\textbf{Step-3: Human annotation} \\
    \hline
    question: 군나르 뮈르달의 주요 경제 모델 중 자본주의와 민주주의의 원칙을 조화시키면서 경제적 평등과 사회적 정의를 추구하는 정치 이념을 뜻하는 것은 무엇인가? \\
    answer: 사회민주주의 \\
    \hline
    question: 군나르 뮈르달의 발전경제학에서 나타나는 주요 특징은 무엇인가? (모두 선택) \\
    choices: [ a) 거시적 지표의 성장, b) 전 사회적인 변화와의 결합, c) 사회공학적 접근, d) 정책 도입 전후의 사회적 영향 분석 ] \\
    answer: [ b, c ] \\
    \hline
    question: 군나르 뮈르달의 발전경제학 접근법에서 강력한 중앙의 권위에 근거한 발전을 상정할 수 밖에 없는 이유는 무엇인가? \\
    choices: [
            a) 사회적 변화의 복잡성,
            b) 사회공학적 접근의 고수,
            c) 근대화 이론에 대한 반대,
            d) 미국 주도의 자본주의 질서
        ] \\
    answer: b \\
    \hline
    question: 군나르 뮈르달의 발전경제학은 단순한 경제 지표의 성장에 초점을 맞추고 있다. (참/거짓) \\
    answer: 거짓 \\
    \hline
    \end{tabular}
    \end{center}
    \caption{Example of step-by-step data generation process for Korean.}
    \label{tab:data_proc_ex_ko}
\end{table*}

\label{appx:data_construction}
\noindent
Tables~\ref{tab:data_proc_ex_en} and \ref{tab:data_proc_ex_ko} show examples of benchmark data at the 1st synthetic, 2nd synthetic, and human annotation stages, as constructed via the pipeline shown in Figure~\ref{fig:cons_pipeline}. The examples present representative questions: short-answer, multiple-choice, and boolean questions in order.

\subsubsection{Short-Answer}
\noindent
We design a multi-stage process for enhancing question quality.
For the initial stage in Table~\ref{tab:data_proc_ex_en}, short-answer questions generated using GPT often suffer from broad phrasing and ambiguous answer candidates. 
For example, a question like ``Which industries have experienced significant advancements due to GPT applications?'' is open to multiple interpretations, making it difficult to derive a single correct answer.
To address this, in the second stage, an automatic refinement process is applied to narrow the scope of the question and include specific clues, enhancing answer steerability.
In this process, by inserting meaning-based clues such as \textit{personalized learning} and \textit{automation of academic tasks}, we enhance the reasoning ability required for the model to understand the context and derive the correct answer.
Finally, through human annotation, we enhance the naturalness and clarity of the question phrasing and further refine the questions by adding higher-level meaning-based clues such as \textit{knowledge transfer}, \textit{intelligent tutoring systems}, and \textit{content generation}.
Through these staged improvements, questions are refined to require semantic inference and contextual understanding rather than simple information retrieval, and are designed to be solvable using LLMs' parametric knowledge without an accompanying paragraph. This contributes to precisely evaluating models' complex language abilities.

Similarly for Korean, as shown in Table~\ref{tab:data_proc_ex_ko}, problems generated in the 1st synthetic data have multiple correct answers. In contrast, when generating the 2nd synthetic data by sampling from the 1st generated data, problems with a single correct answer are generated. This demonstrates that the automatic data generation pipeline model proposed in this paper assists in generating problems with unique answers. Subsequently, through a review process, we enhance the completeness of the problems by adding idioms (or phrases) that enable inference of a single correct answer.

\subsubsection{Multiple Selection}
\noindent
A 3-step process is followed to enhance the quality of multiple-choice questions.
Initial questions of step 1 in Table~\ref{tab:data_proc_ex_en} are broad, such as ``What are some of the main technologies used in GPT models?'', and present general options, making them solvable based solely on superficial information.
Consequently, this results in a limitation where models can achieve high accuracy by relying on simple keyword matching.
In step 2, by including conceptual keywords in the question, such as \textit{its ability to generalize} and \textit{language patterns beyond memorization}, we improve them to require understanding and reasoning about GPT's working principles, rather than simple knowledge retrieval.
Finally, in the human annotation stage, we incorporate higher-level concepts into the questions, such as \textit{hierarchical representations of language}, \textit{contextually relevant responses}, and \textit{semantically coherent}, and also refine the options to subtly distinguish the roles of technical components, elevating the questions to a level where models must understand the function of each component and infer the correct answer.
Through these staged improvements, questions are progressively improved from surface information extraction types to meaning-based inference types, enabling a more refined evaluation of GPT models' complex language understanding and reasoning abilities.

\subsubsection{Multiple Choice} 
\noindent
Consistent with~\cite{joshi-etal-2017-triviaqa}, questions demanding inference capabilities are known to exhibit a higher difficulty than those that do not. As illustrated in Table~\ref{tab:data_proc_ex_ko}, the 1st synthetic data generation stage produces questions focusing on general patterns, such as inquiring about \textit{what} and \textit{major limitations}. In contrast, the 2nd synthetic data stage generates questions that explicitly require inferential reasoning by exploring the interrelations between theoretical approaches and their underlying premises. This demonstrates that providing initial sample data enables the generation of higher-difficulty questions. All generated data undergoes an additional review process to ensure that the answers to the questions are grounded in the original source data and constitute valid responses.

\subsubsection{Boolean} 
For the boolean presented in Table~\ref{tab:data_proc_ex_en},~\ref{tab:data_proc_ex_ko} the 1st synthetic data generation stage produced questions with clear distinctions, largely due to the use of absolute (all-or-nothing) expressions. These questions typically require only a straightforward factual verification to answer. Conversely, the questions generated in the 2nd synthetic data stage posed significant challenges in logical judgment, necessitating a thorough examination of the overall context for resolution. 

Consequently, it can be observed that the iterative question generation method proposed in this paper demands greater knowledge and inference capabilities compared to questions generated via single prompting. During the review process, questions that do not require modification are retained as 2nd-stage data, ensuring the quality of the higher-difficulty set.

\section{Full evaluation results}
\label{appx:full-eval}
\begin{table*}[!t]
    \centering
    \resizebox{\textwidth}{!}{
    \begin{tabular}{lcccccccccc}
    \toprule
    \multirow{2}{*}{\begin{tabular}[c]{@{}l@{}}Model\end{tabular}} & \multicolumn{3}{c}{Summarization}		& {Short Answer}            & \multicolumn{3}{c}{Multiple Selection}	& \multirow{2}{*}{\begin{tabular}[c]{@{}l@{}}MCQ\end{tabular}} & \multirow{2}{*}{\begin{tabular}[c]{@{}l@{}}Boolean\end{tabular}} & \multirow{2}{*}{\begin{tabular}[c]{@{}l@{}}Avg\end{tabular}} \\ \cmidrule(lr){2-4} \cmidrule(lr){5-5} \cmidrule(lr){6-8} 
      & R-1	& R-2	& R-L	 & BERTScore	& A-2	& A-3	& A-4	&    &         &     \\ \hline
    o3-mini  & 0.392 & 0.130 & 0.320 & 0.898 & 0.783 & 0.611 & 0.591 & 0.886 & 0.875 & 0.609 \\
    o1-mini & 0.409 & 0.151 & 0.348 & 0.898 & 0.795 & 0.656 & 0.632 & 0.883 & 0.872 & 0.627 \\
    GPT-4o  & 0.417 & 0.151 & 0.356 & 0.897 & 0.793 & 0.667 & 0.643 & 0.886 & 0.854 & 0.629 \\
    Qwen-72b & 0.396 & 0.151 & 0.345 & 0.902 & 0.746 & 0.650 & 0.628 & 0.900 & 0.896 & 0.624 \\
    llama-70b & 0.397 & 0.146 & 0.341 & 0.903 & 0.751 & 0.620 & 0.598 & 0.875 & 0.873 & 0.612 \\
    Bllossom-70b & 0.349 & 0.129 & 0.299 & 0.880 & 0.676 & 0.528 & 0.507 & 0.767 & 0.840 & 0.553 \\
    Qwen-32b-reasoning & 0.350 & 0.116 & 0.303 & 0.896 & 0.706 & 0.572 & 0.552 & 0.870 & 0.862 & 0.581 \\
    Exaone-32b & 0.321 & 0.094 & 0.267 & 0.889 & 0.463 & 0.611 & 0.583 & 0.876 & 0.879 & 0.554 \\
    Exaone-32b-reasoning & 0.316 & 0.092 & 0.267 & 0.886 & 0.697 & 0.558 & 0.542 & 0.848 & 0.840 & 0.561 \\
    Gemma2-27b & 0.329 & 0.117 & 0.283 & 0.891 & 0.625 & 0.516 & 0.497 & 0.736 & 0.767 & 0.529 \\
    Mistral-24b & 0.414 & 0.159 & 0.359 & 0.901 & 0.756 & 0.580 & 0.560 & 0.865 & 0.827 & 0.602 \\
    Gemma2-9b & 0.294 & 0.096 & 0.248 & 0.883 & 0.577 & 0.471 & 0.449 & 0.695 & 0.761 & 0.497 \\
    Exaone-8b & 0.317 & 0.092 & 0.265 & 0.883 & 0.746 & 0.567 & 0.537 & 0.855 & 0.883 & 0.572 \\
    Mistral-8b & 0.402 & 0.151 & 0.350 & 0.891 & 0.708 & 0.548 & 0.524 & 0.848 & 0.861 & 0.587 \\
    llama-8b & 0.381 & 0.136 & 0.327 & 0.895 & 0.701 & 0.558 & 0.536 & 0.832 & 0.827 & 0.577 \\
    Bllossom-8b & 0.346 & 0.129 & 0.301 & 0.883 & 0.558 & 0.464 & 0.435 & 0.757 & 0.729 & 0.511 \\
    Qwen-7b & 0.388 & 0.144 & 0.338 & 0.896 & 0.713 & 0.573 & 0.550 & 0.844 & 0.856 & 0.589 \\
    \bottomrule 
    \end{tabular}
    }
    \caption{Full table of comprehensibility evaluation results from paragraph-based prompting experiments (Table~\ref{tab:comprehension_eval}).}

    \label{tab:comp-all}
\end{table*}

\begin{table*}[!t]
    \centering
    \resizebox{\textwidth}{!}{
    \begin{tabular}{lcccccccccc}
    \toprule
    \multirow{2}{*}{\begin{tabular}[c]{@{}l@{}}Model\end{tabular}} & \multicolumn{3}{c}{Summarization}		& {Short Answer}            & \multicolumn{3}{c}{Multiple Selection}	& \multirow{2}{*}{\begin{tabular}[c]{@{}l@{}}MCQ\end{tabular}} & \multirow{2}{*}{\begin{tabular}[c]{@{}l@{}}Boolean\end{tabular}} & \multirow{2}{*}{\begin{tabular}[c]{@{}l@{}}Avg\end{tabular}} \\ \cmidrule(lr){2-4} \cmidrule(lr){5-5} \cmidrule(lr){6-8} 
      & R-1	& R-2	& R-L	 & BERTScore	& A-2	& A-3	& A-4	&    &         &     \\ \hline
        o3-mini  & 0.391 & 0.130 & 0.321 & 0.898 & 0.780 & 0.612 & 0.593 & 0.886 & 0.876 & 0.610 \\
        o1-mini & 0.409 & 0.151 & 0.348 & 0.897 & 0.786 & 0.651 & 0.628 & 0.885 & 0.874 & 0.626 \\
        GPT-4o  & 0.414 & 0.149 & 0.352 & 0.898 & 0.796 & 0.669 & 0.646 & 0.880 & 0.847 & 0.628 \\
        Qwen-72b & 0.317 & 0.116 & 0.278 & 0.896 & 0.741 & 0.634 & 0.613 & 0.885 & 0.886 & 0.596 \\
        llama-70b & 0.383 & 0.141 & 0.336 & 0.902 & 0.731 & 0.619 & 0.596 & 0.881 & 0.875 & 0.607 \\
        Bllossom-70b & 0.220 & 0.079 & 0.192 & 0.849 & 0.548 & 0.423 & 0.406 & 0.628 & 0.739 & 0.454 \\
        Exaone-32b & 0.201 & 0.059 & 0.169 & 0.815 & 0.721 & 0.570 & 0.548 & 0.787 & 0.661 & 0.503 \\
        Gemma2-27b & 0.137 & 0.038 & 0.119 & 0.833 & 0.481 & 0.404 & 0.391 & 0.523 & 0.617 & 0.394 \\
        Mistral-24b & 0.358 & 0.142 & 0.315 & 0.895 & 0.714 & 0.580 & 0.561 & 0.843 & 0.834 & 0.582 \\
        Gemma2-9b & 0.090 & 0.019 & 0.082 & 0.834 & 0.467 & 0.394 & 0.379 & 0.539 & 0.604 & 0.379 \\
        Exaone-8b & 0.164 & 0.045 & 0.141 & 0.556 & 0.541 & 0.414 & 0.392 & 0.804 & 0.827 & 0.432 \\
        Mistral-8b & 0.324 & 0.119 & 0.283 & 0.882 & 0.580 & 0.498 & 0.476 & 0.821 & 0.822 & 0.534 \\
        llama-8b & 0.368 & 0.132 & 0.322 & 0.872 & 0.553 & 0.464 & 0.435 & 0.618 & 0.794 & 0.507 \\
        Bllossom-8b & 0.181 & 0.064 & 0.159 & 0.873 & 0.367 & 0.313 & 0.300 & 0.611 & 0.651 & 0.391 \\
        Qwen-7b & 0.252 & 0.082 & 0.215 & 0.864 & 0.640 & 0.501 & 0.480 & 0.793 & 0.836 & 0.518 \\
    \bottomrule 
    \end{tabular}
    }
    \caption{Overall evaluation results for paragraph w/ CoT from Table~\ref{tab:comp-all}.}
    \label{tab:para-cot-all}
\end{table*}

\begin{table*}[!t]
    \centering
    \resizebox{\textwidth}{!}{
    \begin{tabular}{lcccccccccc}
    \toprule
    \multirow{2}{*}{\begin{tabular}[c]{@{}l@{}}Model\end{tabular}} & \multicolumn{3}{c}{Summarization}		& {Short Answer}            & \multicolumn{3}{c}{Multiple Selection}	& \multirow{2}{*}{\begin{tabular}[c]{@{}l@{}}MCQ\end{tabular}} & \multirow{2}{*}{\begin{tabular}[c]{@{}l@{}}Boolean\end{tabular}} & \multirow{2}{*}{\begin{tabular}[c]{@{}l@{}}Avg\end{tabular}} \\ \cmidrule(lr){2-4} \cmidrule(lr){5-5} \cmidrule(lr){6-8} 
      & R-1	& R-2	& R-L	 & BERTScore	& A-2	& A-3	& A-4	&    &         &     \\ \hline
    o3-mini  & 0.392 & 0.130 & 0.320 & 0.898 & 0.783 & 0.611 & 0.591 & 0.886 & 0.875 & 0.609 \\
    o1-mini & 0.409 & 0.151 & 0.348 & 0.898 & 0.795 & 0.656 & 0.632 & 0.883 & 0.872 & 0.627 \\
    GPT-4o  & 0.417 & 0.151 & 0.356 & 0.897 & 0.793 & 0.667 & 0.643 & 0.886 & 0.854 & 0.629 \\
    Qwen-72b & 0.396 & 0.151 & 0.345 & 0.902 & 0.746 & 0.650 & 0.628 & 0.900 & 0.896 & 0.624 \\
    llama-70b & 0.397 & 0.146 & 0.341 & 0.903 & 0.751 & 0.620 & 0.598 & 0.875 & 0.873 & 0.612 \\
    Bllossom-70b & 0.349 & 0.129 & 0.299 & 0.880 & 0.676 & 0.528 & 0.507 & 0.767 & 0.840 & 0.553 \\
    Qwen-32b-reasoning & 0.350 & 0.116 & 0.303 & 0.896 & 0.706 & 0.572 & 0.552 & 0.870 & 0.862 & 0.581 \\
    Exaone-32b & 0.321 & 0.094 & 0.267 & 0.889 & 0.463 & 0.611 & 0.583 & 0.876 & 0.879 & 0.554 \\
    Exaone-32b-reasoning & 0.316 & 0.092 & 0.267 & 0.886 & 0.697 & 0.558 & 0.542 & 0.848 & 0.840 & 0.561 \\
    Gemma2-27b & 0.329 & 0.117 & 0.283 & 0.891 & 0.625 & 0.516 & 0.497 & 0.736 & 0.767 & 0.529 \\
    Mistral-24b & 0.414 & 0.159 & 0.359 & 0.901 & 0.756 & 0.580 & 0.560 & 0.865 & 0.827 & 0.602 \\
    Gemma2-9b & 0.294 & 0.096 & 0.248 & 0.883 & 0.577 & 0.471 & 0.449 & 0.695 & 0.761 & 0.497 \\
    Exaone-8b & 0.317 & 0.092 & 0.265 & 0.883 & 0.746 & 0.567 & 0.537 & 0.855 & 0.883 & 0.572 \\
    Mistral-8b & 0.402 & 0.151 & 0.350 & 0.891 & 0.708 & 0.548 & 0.524 & 0.848 & 0.861 & 0.587 \\
    llama-8b & 0.381 & 0.136 & 0.327 & 0.895 & 0.701 & 0.558 & 0.536 & 0.832 & 0.827 & 0.577 \\
    Bllossom-8b & 0.346 & 0.129 & 0.301 & 0.883 & 0.558 & 0.464 & 0.435 & 0.757 & 0.729 & 0.511 \\
    Qwen-7b & 0.388 & 0.144 & 0.338 & 0.896 & 0.713 & 0.573 & 0.550 & 0.844 & 0.856 & 0.589 \\
    \bottomrule 
    \end{tabular}
    }
    \caption{Overall evaluation results for English based on Table~\ref{tab:main-result}.}
    \label{tab:comp-all-main-en}
\end{table*}

\begin{table*}[!t]
    \centering
    \resizebox{\textwidth}{!}{
    \begin{tabular}{lcccccccccc}
    \toprule
    \multirow{2}{*}{\begin{tabular}[c]{@{}l@{}}Model\end{tabular}} & \multicolumn{3}{c}{Summarization}		& {Short Answer}            & \multicolumn{3}{c}{Multiple Selection}	& \multirow{2}{*}{\begin{tabular}[c]{@{}l@{}}MCQ\end{tabular}} & \multirow{2}{*}{\begin{tabular}[c]{@{}l@{}}Boolean\end{tabular}} & \multirow{2}{*}{\begin{tabular}[c]{@{}l@{}}Avg\end{tabular}} \\ \cmidrule(lr){2-4} \cmidrule(lr){5-5} \cmidrule(lr){6-8} 
      & R-1	& R-2	& R-L	 & BERTScore	& A-2	& A-3	& A-4	&    &         &     \\ \hline
    o1-mini & 0.482 & 0.206 & 0.406 & 0.863 & 0.521 & 0.434 & 0.402 & 0.683 & 0.747 & 0.527 \\
    o3-mini  & 0.466 & 0.180 & 0.370 & 0.868 & 0.612 & 0.497 & 0.464 & 0.720 & 0.756 & 0.548 \\
    GPT-4o  & 0.493 & 0.206 & 0.416 & 0.864 & 0.556 & 0.498 & 0.469 & 0.739 & 0.694 & 0.548 \\
    Qwen-72b & 0.472 & 0.209 & 0.409 & 0.867 & 0.461 & 0.424 & 0.408 & 0.752 & 0.825 & 0.536 \\
    llama-70b & 0.457 & 0.193 & 0.389 & 0.866 & 0.526 & 0.465 & 0.436 & 0.752 & 0.798 & 0.542 \\
    Bllossom-70b & 0.465 & 0.197 & 0.389 & 0.847 & 0.695 & 0.417 & 0.406 & 0.642 & 0.787 & 0.538 \\
    Qwen-32b-reasoning & 0.430 & 0.166 & 0.368 & 0.861 & 0.545 & 0.454 & 0.419 & 0.697 & 0.777 & 0.524 \\
    Exaone-32b & 0.411 & 0.147 & 0.343 & 0.861 & 0.548 & 0.445 & 0.414 & 0.716 & 0.713 & 0.511 \\
    Exaone-32b-reasoning & 0.368 & 0.128 & 0.313 & 0.848 & 0.477 & 0.384 & 0.356 & 0.680 & 0.575 & 0.459 \\
    Gemma2-27b & 0.442 & 0.183 & 0.375 & 0.864 & 0.523 & 0.443 & 0.418 & 0.710 & 0.827 & 0.532 \\
    Mistral-24b & 0.488 & 0.214 & 0.420 & 0.859 & 0.533 & 0.461 & 0.436 & 0.717 & 0.611 & 0.526 \\
    Gemma2-9b & 0.412 & 0.165 & 0.342 & 0.855 & 0.493 & 0.426 & 0.401 & 0.670 & 0.796 & 0.507 \\
    llama-8b & 0.440 & 0.181 & 0.372 & 0.845 & 0.421 & 0.368 & 0.346 & 0.657 & 0.485 & 0.457 \\
    Mistral-8b & 0.472 & 0.204 & 0.407 & 0.844 & 0.426 & 0.356 & 0.343 & 0.628 & 0.610 & 0.477 \\
    Exaone-8b & 0.393 & 0.138 & 0.330 & 0.852 & 0.554 & 0.426 & 0.402 & 0.687 & 0.735 & 0.502 \\
    Bllossom-8b & 0.468 & 0.200 & 0.402 & 0.843 & 0.479 & 0.390 & 0.354 & 0.616 & 0.481 & 0.470 \\
    Qwen-7b & 0.472 & 0.203 & 0.407 & 0.848 & 0.512 & 0.440 & 0.412 & 0.685 & 0.722 & 0.522 \\
    \bottomrule 
    \end{tabular}
    }
    \caption{Overall evaluation results for Korean based on Table~\ref{tab:main-result}.}
    \label{tab:comp-all-main-ko}
\end{table*}

\subsection{Evaluation results for comprehensibility}
\noindent
Table~\ref{tab:comp-all} presents the evaluation results for comprehensibility across all models using paragraph-augmented prompting. This setting provides each model with an extended input that includes contextual paragraphs to assess its ability to understand and interpret academic content more effectively.
\subsection{Evaluation results for comprehensibility with CoT}
\noindent
Table~\ref{tab:para-cot-all} shows the results of the comprehensibility evaluation when Chain-of-Thought (CoT) prompting was applied. This experimental setting prompts models to generate intermediate reasoning steps before producing a final response, aiming to enhance interpretability and answer quality.
\subsection{Overall evaluation results for English}
\noindent
Table~\ref{tab:comp-all-main-en} summarizes the English-only evaluation results extracted from the full paragraph-augmented prompting experiments. The results reflect the models’ performance specifically on English inputs across all academic domains and question types.
\subsection{Overall evaluation results for Korean}
\noindent
Table~\ref{tab:comp-all-main-ko} reports the evaluation results for Korean-language inputs, also based on the full paragraph-augmented prompting experiments. This analysis focuses on assessing multilingual capability by isolating performance on Korean prompts.

\section{Qualitative Results}
\noindent
\begin{table*}[t]
\centering
\small
\resizebox{\textwidth}{!}{
\begin{tabular}{p{0.1\linewidth} p{0.4\linewidth} p{0.38\linewidth} p{0.2\linewidth}}
\toprule
\textbf{Category} & \textbf{Paragraph (excerpt)} & \textbf{Question} & \textbf{Answer}  \\
\midrule
\multirow{4}{*}{\parbox{\linewidth}{Economy \& Management}} &
\multirow{4}{*}{
\parbox{\linewidth}
{In modern organizations, most forms of overt gender discrimination (i.e., blatant mistreatment or overtly sexistjokes) have become less socially acceptable and have beenreplaced with subtle and often unintentional slights, knownas microaggressions that denigrate women (Capodilupoet al., 2010; Cardador, 2017; Cortina et al., 2013; Yang \& Carroll, 2018). To illustrate, Tracy Chou, an experienced software engineer, \\
... \\
work strategies women use, and the buffers that influence their sensemaking process.
}}
& What is the term for comments that subtly and unintentionally denigrate women's competence in professional settings? & Subtle discriminatory comments  \\ \cmidrule{3-4}
& & Which of the following are types of gender microaggressions encountered by women in STEM? (Select all that apply) & a) Microassault,
                b) Microinsult,
                c) Microinvalidation \\ \cmidrule{3-4}
& & What is one effect of microaggressions on women in STEM? &         c) Negative psychological outcomes  \\ \cmidrule{3-4}
& & Microaggressions may negatively influence a woman's professional identity. (True/False) & True \\
\midrule
\multirow{4}{*}{\parbox{\linewidth}{Chemical \& Biochemistry}} &
\multirow{4}{*}{\parbox{\linewidth}{
The average size from Cryo-TEM was around 57\% smaller than that from SN-FSHS-CICS, which might be partially attributed to the physical difference in the size characterized: Under CryoTEM, only the electron-dense region, presumably the core ensemble of lipids and RNA, is captured, \\
... \\
Subsequently, the correlation between each fluorescently tagged payload and LNP size was better visualized by projecting the 3D data onto the corresponding planes for the Cy3-siRNA payload (Figure 4d) \\
...
}}
& How does increasing PEG lipid content in LNPs affect the siRNA payload? & Increased PEG decreases siRNA payload  \\ \cmidrule{3-4}
& & Which techniques were used for data analysis of siRNA LNPs size and loading? (Select all that apply) & a) SN-FSHS-CICS,
                b) Cryo-TEM \\ \cmidrule{3-4}
    & & How does the average siRNA payload per LNP change with PEG concentration? &    c) It decreases with PEG concentration.  \\ \cmidrule{3-4}
& & The increasing percentage of PEG in formulations leads to larger average sizes of LNPs. (True/False) & False \\
\midrule
\multirow{4}{*}{\parbox{\linewidth}{Engineering}} &
\multirow{4}{*}{\parbox{\linewidth}{
 It’s essential to recognize that GPTs might occasionally make mistakes or give poor answers, particularly when dealing with complicated or ambiguous queries. This highlights the necessity of continual model training, thorough testing, and modification to guarantee that they consistently meet consumer needs. To confirm the efficacy and dependability of using GPTs specifically in the e-commerce area, more research and testing are required.  \\
...
}}
& What is the main aspect HCI addresses in terms of GPT usability? & User interaction efficiency  \\ \cmidrule{3-4}
& & Which ethical issues are related to GPT models? (Select all that apply) & a) Privacy concerns, b) Data bias \\ \cmidrule{3-4}
    & & What is a potential disadvantage of HCI in GPT models? &   b) Potential for biases  \\ \cmidrule{3-4}
& & HCI techniques improve GPT usability but might introduce biases. (True/False) & True \\
\midrule
\multirow{4}{*}{\parbox{\linewidth}{Medical Science}} &
\multirow{4}{*}{\parbox{\linewidth}{
our results showed that targeting all three subpopulations with 4-1BB activation and not only the stem-like T cells with OX40 activation endowed HBV-specific CD8[+] T cells with robust antiviral activity. The mechanism behind this observation remains uncertain and may be linked to lower TSL numbers, their potential distinct localization, or the differential biological effects downstream of these two co-stimulatory receptors.[37][,][38]  The potential of 4-1BB agonism for initiating anti-tumor T cell responses is well recognized. \\
...  \\
}}
& What therapeutic target is known to reinvigorate dysfunctional HBV-specific CD8[+] T cells? & 4-1BB  \\ \cmidrule{3-4}
& & What factors affect the proliferation of CD8[+] T cells in the context of dysregulation? (Select all that apply) & a) Co-stimulation, b) Cytokine environment, c) Ag engagement \\ \cmidrule{3-4}
    & & Which molecule is expressed exclusively by the Dys-TSL population? &   b) OX40  \\ \cmidrule{3-4}
& & The activation of OX40 leads to a significant increase in CD8[+] T cells' ability to produce IFN-g. (True/False) & False \\
\midrule
\multirow{4}{*}{\parbox{\linewidth}{Biology \& Earth Science}} &
\multirow{4}{*}{\parbox{\linewidth}{
... provided in Table S3. Upon analysing the dyeing performance depicted in Fig. 3A, it be  comes clear that both the wool fibers with and without mordant exhibited comparable chlorophyll uptakes, with the unmordanted wool fibers even demonstrating higher chlorophyll uptake values. These results are particularly intriguing, as the fixation of natural dyes in textile fibers typically requires mordanting processes prior to the dyeing cycle in order to enhance the dye uptake (Guesmi et al., 2013; Zhao et al., 2020a). Moreover, achieving a natural dye uptake exceeding 70\% without any type of optimization ... 
}}
& What solvent is primarily used in the ABS process to extract chlorophyll?  & Ethanol  \\ \cmidrule{3-4}
& & Which components contribute to the recovery process in ABS? (Select all that apply) & a) CuSO4, b) Chlorophyll derivatives, c) Sodium hydroxide, d) Ethanol \\ \cmidrule{3-4}
    & & What was observed regarding dye uptake in unmordanted fibers compared to mordanted fibers? &  b) Unmordanted fibers showed equal or greater dye uptake.  \\ \cmidrule{3-4}
& & The ABS process in the study was shown to have potential health risks associated with pollution. (True/False) & True \\
\bottomrule
\end{tabular}
}
\caption{Qualitative results for each topic category I.}
\label{tab:qual_res}
\end{table*}

\begin{table*}[t]
\centering
\small
\resizebox{\textwidth}{!}{
\begin{tabular}{p{0.1\linewidth} p{0.4\linewidth} p{0.38\linewidth} p{0.2\linewidth}}
\toprule
\textbf{Category} & \textbf{Paragraph (excerpt)} & \textbf{Question} & \textbf{Answer}  \\
\midrule
\multirow{4}{*}{\parbox{\linewidth}{Physics \& Mathematics}} &
\multirow{4}{*}{\parbox{\linewidth}{
... We then show posterior distributions obtained, respectively, with runs adopting 3k, 6k, and 10k live points, demonstrating the gradual convergence to slightly lower values of radii and larger uncertainties...\\
... the main mode identified with the ST-U model and reported in panel (A) of the same Figure. The omitting component is always associated with the smaller, closer-to-the-equator, hot spot (labeled as primary in panel (A) of Figure 3). The location and size of the masking element can vary significantly within the identified mode... \\
... The omitting component is always associated with the smaller, closer-to-the-equator, hot spot (labeled as primary in panel (A) of Figure 3). The location and size of the masking element can vary significantly within the identified mode \\
...
}}
& The NICER instrument collected 1.936 Ms (megaseconds) of data from PSR J0030$+$0451 over a specific time period. Convert this time into days. (1 Ms = $10^(6)$ seconds)  & 22.4 days  \\ \cmidrule{3-4}
& & What components are included in the ST+PST model? (Select all that apply) & a)Primary hot spot, b)Secondary hot spot emitting, c)Secondary hot spot masking \\ \cmidrule{3-4}
    & & The data for PSR J0030+0451 included multiple inference runs with various live point (LP) settings. If one inference run used 10,000 live points and another used half that amount, how many live points did the second run use? &  c) 5,000 \\ \cmidrule{3-4}
& & Multimodal structures in a posterior surface suggest the existence of multiple solutions or interpretations for a model's parameters. (True/False) & True \\
\midrule
\multirow{4}{*}{\parbox{\linewidth}{Social Science}} &
\multirow{4}{*}{\parbox{\linewidth}{
 When management seeks to incorporate all stakeholders and their demands, this  may only be feasible in a sequential manner where some stakeholders lose with regard to some aspects in the short term and others gain,  but then sequential negotiations can help creating packages that foster sustainable development at a societal and planetary level. \\
... \\
Moreover, Tesla's success has produced a wave  of start-ups across the world vying to make EVs at a lower cost than  Tesla can. While EVs are not a perfect solution, and Tesla is not a perfect ... 
}}
& How do firms engage in sustainable entrepreneurship by working with others?  & Collaborative innovation  \\ \cmidrule{3-4}
& &  What actions are essential to enhance sustainability? (Select all that apply) & a) Co-creating policies, c) Engaging in responsible lobbying  \\ \cmidrule{3-4}
    & & Which action is essential for improving sustainability in corporate practices? &  c) Engaging in stakeholder collaboration  \\ \cmidrule{3-4}
& & Firms can solve complex sustainability issues solely on their own without any external collaboration. (True/False) & False \\
\midrule
\multirow{4}{*}{\parbox{\linewidth}{Humanities, Literature \& Arts}} &
\multirow{4}{*}{\parbox{\linewidth}{
Navigational capital refers to the ability to maneuver through institutions created to exclude groups or classes of people (i.e. the Dominican education system, which both symbolically and physically excludes people of Haitian descent). Social capital refers to people and relationships that provide emotional and instrumental support when navigating systems, like schools and government bureaucracies. Linguistic capital includes the cognitive flexibility and social skills that come with the ability to navigate multiple languages. Familial capital involves the history, memory and cultural intuition that one gains through an extended  ...
}}
& What specific challenge related to documentation impacts school participation for Dominican females of Haitian descent?  & Lack of documentation  \\ \cmidrule{3-4}
& &  What factors contribute to the educational challenges faced by Dominican girls of Haitian descent? (Select all that apply) & a) Cultural attitudes like machismo, b) Economic hardship  \\ \cmidrule{3-4}
    & & What impact can the lack of documentation have on youth education? &  a) Denial of access to national exams  \\ \cmidrule{3-4}
& & The absence of documentation does not affect the educational success of Dominican females of Haitian descent. (True/False) & False \\
\bottomrule
\end{tabular}
}
\caption{Qualitative results for each topic category II.}
\label{tab:qual_res2}
\end{table*}

Table~\ref{tab:qual_res} shows qualitative results. 

\end{document}